\pdfoutput=1
\documentclass[11pt]{article}
\usepackage{tabularx} % 用于自适应宽度的表格
\usepackage[table]{xcolor}
\usepackage{acl}

\usepackage{times}
\usepackage{latexsym}
\usepackage{booktabs}
\usepackage{xcolor}
\definecolor{mydarkgreen}{rgb}{0.0, 0.5, 0.0}
\newcommand{\mydarkgreen}[1]{{\color{mydarkgreen}#1}}
\usepackage{amssymb}
\usepackage[T1]{fontenc}
\usepackage[utf8]{inputenc}
\usepackage{algorithm}
\usepackage{microtype}
\usepackage{makecell}
\usepackage{pifont}
\usepackage{arydshln}
\usepackage{multirow}
\usepackage{ulem}
\usepackage{multicol}
\usepackage{algpseudocode}  
\usepackage{amsmath,amssymb,amsfonts}
\usepackage{subfigure}
\usepackage{marvosym}
\usepackage{amsthm}
\usepackage{pgfplots}
\definecolor{mygray}{gray}{0.8}
\usepackage{tabularx}
\pgfplotsset{compat=1.16}
\usepackage{newfloat}
\usepackage{listings}
\usepackage{CJK}
\usepackage{microtype}

\usepackage{inconsolata}

\usepackage{color}

\usepackage{colortbl}
\definecolor{mygray}{gray}{0.8}
\newcommand{\red}[1]{{\color{red}{#1}}}
\newcommand{\black}[1]{{\color{black}{#1}}}

\newcommand{\myred}{\black}

\newcommand{\cdashlinelr}[1]{%
  \noalign{\vskip\aboverulesep
           \global\let\@dashdrawstore\adl@draw
           \global\let\adl@draw\adl@drawiv}
  \cdashline{#1}
  \noalign{\global\let\adl@draw\@dashdrawstore
           \vskip\belowrulesep}}

\definecolor{lightblue}{RGB}{173, 216, 230}
\definecolor{lightyellow}{RGB}{255, 255, 224}
\definecolor{lightred}{RGB}{255, 182, 193}
\definecolor{lightpurple}{RGB}{216, 191, 216}

\definecolor{darkblue}{RGB}{123, 166, 180}   
\definecolor{darkyellow}{RGB}{245, 245, 184} 
\definecolor{darkred}{RGB}{205, 132, 143}    
\definecolor{darkpurple}{RGB}{166, 141, 166} 
\title{DetectBench: Can Large Language Model Detect and Piece Together Implicit Evidence?}

\author{Zhouhong Gu\textsuperscript{\rm $\spadesuit$}\thanks{\ \ Equal Contribution}\ ,
Lin Zhang\textsuperscript{\rm $\spadesuit$\ *},
Xiaoxuan Zhu\textsuperscript{\rm $\spadesuit$},
Jiangjie Chen\textsuperscript{\rm $\spadesuit$},
Wenhao Huang\textsuperscript{\rm $\spadesuit$},\\
\bf
Yikai Zhang\textsuperscript{\rm $\spadesuit$},
Shusen Wang\textsuperscript{\rm $\clubsuit$},
Zheyu Ye\textsuperscript{\rm $\heartsuit$},
Yan Gao\textsuperscript{\rm $\heartsuit$},
Yanghua Xiao\textsuperscript{\rm $\spadesuit$}\thanks{\ \ Corresponding authors.}\ ,
Hongwei Feng\textsuperscript{\rm $\spadesuit$}\footnotemark[2]
\\
\textsuperscript{\rm $\spadesuit$}Shanghai Key Laboratory of Data Science, School of Computer Science, Fudan University\\
\textsuperscript{\rm $\clubsuit$}Meta\quad
\textsuperscript{\rm $\heartsuit$}Xiaohongshu Inc.\\
\texttt{\{zhgu22, linzhang22, xxzhu22\}@m.fudan.edu.cn},\\
\texttt{\{zheyuye, yadun\}@xiaohongshu.com},
\texttt{wangshusen@meta.com},\\
\texttt{\{jjchen19, shawyh, hwfeng\}@fudan.edu.cn}}

\begin{document}
\begin{CJK}{UTF8}{gbsn}

\maketitle

\begin{abstract}
Detecting evidence within the context is a key step in the process of reasoning task.
Evaluating and enhancing the capabilities of LLMs in evidence detection will strengthen context-based reasoning performance.
This paper proposes a benchmark called DetectBench for verifying the ability to detect and piece together implicit evidence within a long context.
DetectBench contains 3,928 multiple-choice questions, with an average of 994 tokens per question. 
Each question contains an average of 4.55 pieces of implicit evidence, and solving the problem typically requires 7.62 logical jumps to find the correct answer.
To enhance the performance of LLMs in evidence detection, this paper proposes Detective Reasoning Prompt and Finetune.
Experiments demonstrate that the existing LLMs' abilities to detect evidence in long contexts are far inferior to humans.
However, the Detective Reasoning Prompt effectively enhances the capability of powerful LLMs in evidence detection, while the Finetuning method shows significant effects in enhancing the performance of weaker LLMs.
Moreover, when the abilities of LLMs in evidence detection are improved, their final reasoning performance is also enhanced accordingly.
The benchmark is available in \url{https://github.com/MikeGu721/DetectBench}.

\end{abstract}

\section{Introduction}
\label{sec:intro}
\begin{figure}[t]
    \centering
    \resizebox{\columnwidth}{!}{
    \includegraphics{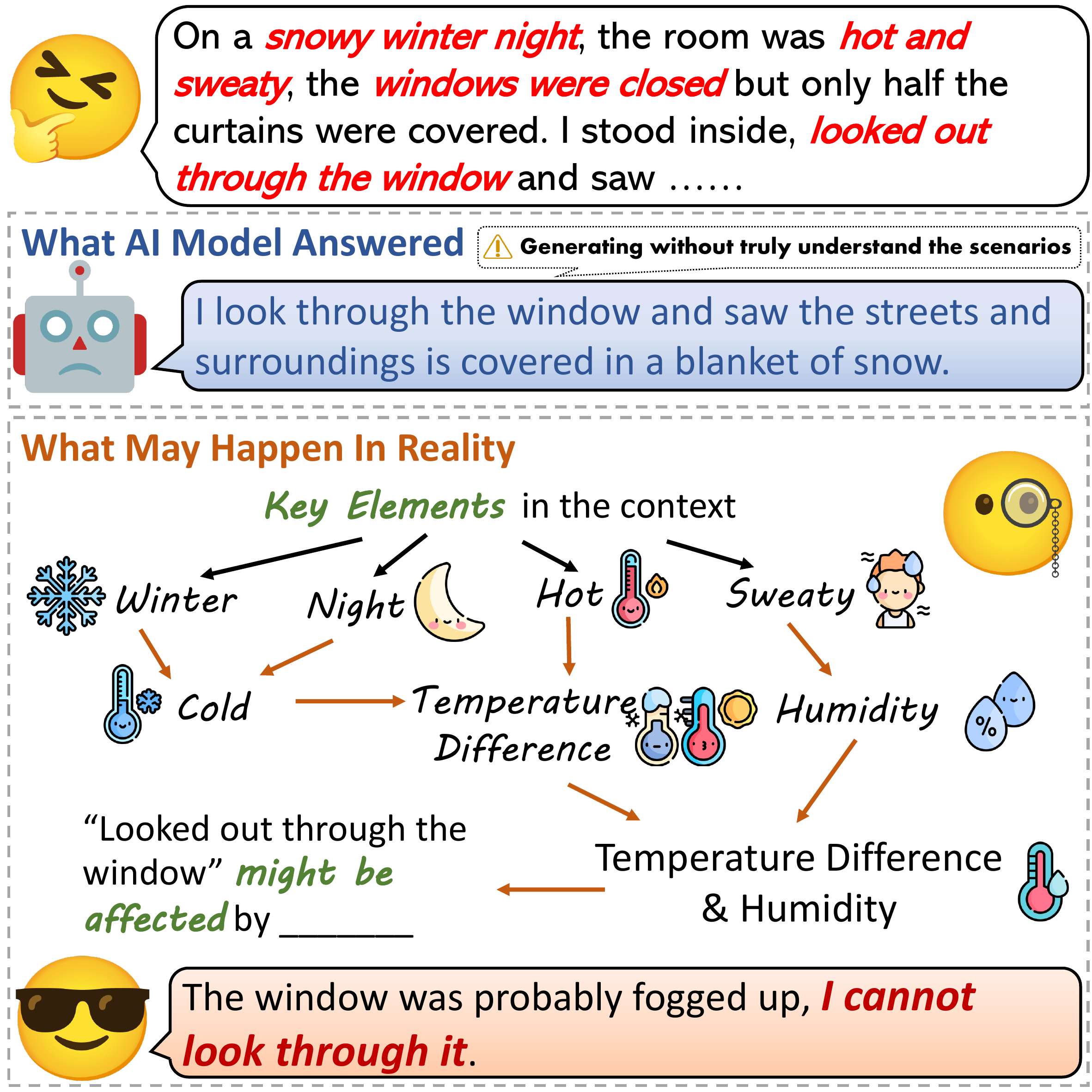}
    }
    \caption{
    LLMs are hard to aware of the implicit evidence in the context so they may respond arbitrarily.
    }
    \label{fig:intro}
% \vspace{-7mm}
\end{figure}

\begin{table*}[t]
\small
    \centering
\resizebox{\textwidth}{!}{
\begin{tabular}{>{\raggedright\arraybackslash}m{2.5cm}>{\raggedright\arraybackslash}m{4cm}>{\raggedright\arraybackslash}m{2cm}>{\raggedright\arraybackslash}m{10cm}}
\toprule
\textbf{Source} & \textbf{Question} & \textbf{How to detect evidence} & \textbf{Context} \\ \hline
SQuAD~\cite{rajpurkar2016squad} & To whom did the \textbf{Virgin Mary} allegedly appear \textbf{in 1858} in Lourdes France? & String Matching & ...... It is a replica of the grotto at Lourdes, France where the \textbf{Virgin Mary} reputedly appeared to Saint Bernadette Soubirous \textbf{in 1858}. At the end of the main drive (and in a direct line that connects through 3 statues and the Gold Dome), is a simple, modern stone statue of Mary.\\ \hline
WikiNLDB~\cite{thorne2021database} & Who studied at \textbf{University of Minnesota}? & String Matching & 1. Melvin Maas graduated from the \textbf{University of Minnesota} and is buried at Arlington National Cemetery. He is a native of Minnesota and his language is English. 2. Clarence Larson graduated from the \textbf{University of Minnesota} and is a member of the National Academy of Engineering..... \\ \hline
WikiNLDB~\cite{thorne2021database} & What is the largest \textbf{yearly} attendance? & Semantic Matching & 1. The Musee en herbe has a visitor \textbf{per year} of 70000. 2. [The total number of visitors to the Hirschsprung Collection is 71779 \textbf{per year}. ... 24. The Tate Modern has a visitor account of 5839197 visitors \textbf{per year}. 25. Catoctin Mountain Park attracts 221750 visitors \textbf{per year}. \\ \hline
HotpotQA~\cite{yang2018hotpotqa} \& LongBench~\cite{bai2023longbench} & Which team does the player \textbf{2015 Diamond Head Classic's MVP} \underline{play for}? & String \& Semantic Matching  & The \textbf{2015 Diamond Head Classic} was a college basketball tournament ... Buddy Hield was named the tournament's \textbf{MVP}...... Chavano Rainier ``Buddy'' Hield is a Bahamian professional basketball \underline{player for} the Sacramento Kings of the NBA...... \\ \hline
StructBench  \textbf{(Ours)} & Is this young man guilty or not? & Deep Understanding To The Context & 
\textbf{On a wintry winter night}, a tragic event unfolded at 68 King's West Road. A single woman was found murdered at the doorstep of her room around 8pm. The scene was set \textbf{in a quietly cozy room, warmed by a gas stove that glowed red-hot}, offering a stark contrast to the cold white blanket enveloping the outside world......\\ 
\bottomrule
\end{tabular}}
    \caption{Difference in the implicity of the evidence between benchmarks that needs to do evidence detection and reasoning combinely.}
    \label{tab:comparsion}
\vspace{-5mm}
\end{table*}

The ability to perform reasoning over natural language is an important aspect of intelligence~\cite{chen2022harnessing}.
Tasks designed to assess inferential capabilities commonly consist of a context and a question, expecting the Large Language Models~(LLMs) to respond correctly~\cite{chu2023survey, davis2023benchmarks}.
Human annotators often conceal the evidence necessary for answering the question within the context.
This raises a question:
\textbf{\textit{whether LLMs possess the capability to detect these pieces of evidence and understand how to formulate reasoning based upon them?}}

Identifying evidence often poses a more significant challenge than reasoning, as it necessitates a deeper understanding of the question and context~\cite{hu2024rag, ding2024survey}.
There are many existing tasks evaluate the model's joint abilities in evidence detection and evidence-based reasoning in long contexts, such as reading comprehension~\cite{yu2020reclor,kazi2021uquad1,lu2022contextual}, retrieval reasoning~\cite{yang2018hotpotqa,chen2023benchmarking}, and fact verification~\cite{thorne2018fever,Thorne19FEVER2,Aly21Feverous}.
As shown in Tab.~\ref{tab:comparsion}, the existing benchmarks~\cite{rajpurkar2016squad,yang2018hotpotqa,thorne2021database,bai2023longbench} of these tasks often present evidence that is too explicit and direct, which is easy to find through rule-based retrieval methods.
However, in real scenarios, evidence is usually implicit within the context, and accurately solving a problem often requires the integration of multiple pieces of evidence through joint reasoning.
For example, as shown in Fig.~\ref{fig:intro}, only when we realize that changes in temperature and humidity will make glass foggy can we figure out that details about temperature and humidity are crucial to seeing through the glass.

To evaluate whether models can detect and piece pieces of evidence together to answer questions, a benchmark consisting of multiple pieces of implicit evidence within a long context is needed.
So, in this paper, we propose a multiple-choice question answering benchmark called \textbf{Detect}ive \textbf{Bench}mark~(\textbf{DetectBench}).
DetectBench comes from the idea that ``when facing a criminal case, detectives often need to identify the most crucial evidence from a vast array of seemingly unrelated information to solve the case''.
This benchmark comprises 3,928 questions, each paired with a paragraph averaging 994 tokens and averaging 4.55 annotated implicit evidence to answer a question.
To delineate the distinctions among high-capacity LLMs, we further introduce DetectBench-Hard and DetectBench-Distract.
DetectBench-Hard comprises the 300 most lengthy and complexly annotated samples from DetectBench.
And DetectBench-Distract consists of 300 manually meticulously expanded questions, averaging 10,779 tokens each, designed to better suit the requirements of LLMs for understanding long context.
The characteristics of DetectBench include:
1. Evidence related to question-answering cannot be detected through the character or string within questions and options.
2. It necessitates the combination of multiple pieces of evidence to derive more critical results for question answering.
% 3. The context contains a significant amount of misleading and irrelevant information.
3. Each question has a detailed manual annotation from evidence to reasoning process and to answer.

In experiments conducted on human participants and LLMs, we assessed their evidence detection and question-answering abilities on DetectBench.
Our findings reveal that humans significantly surpassed the most advanced LLMs in both tasks.
By analyzing the correlation between accuracy in evidence detection and question answering, we discovered a high degree of positive correlation between them, confirming the effectiveness of the annotations within DetectBench and underscoring the critical role of evidence identification in the reasoning process.

To enhance the model's capabilities in evidence detection and evidence-based reasoning, we proposed Detective Reasoning to improve these two capacities simultaneously.
Like how experienced detectives collectively conduct evidence detection and reasoning, Detective Reasoning enhances LLMs by directing them to thoroughly consider all possible evidence, engage in reasoning, and summarize the entire reasoning process to refine the evidence. 
Finally, reasoning from the evidence is used to ascertain the answer to the question.
Constructing prompts with Detective Reasoning further enhances the evidence detection and reasoning capabilities of state-of-the-art~(SoTA) LLMs.
Similarly, developing a Fine-Tuning~(FT) dataset inspired by the principles of Detective Reasoning also advances the abilities of open-source LLMs in this regard.

\section{Related Works}
\label{sec:rel}

\subsection{Information Retrieval}

Evidence detection is one of the two main characteristics of DetectBench, which is a sub-domain of Information Retrieval.
Information Retrieval aims to address pertinent tasks by extracting crucial data from many references, where the most significant challenge lies in identifying implicit key information~\cite{zhu2023large,yang2022survey}. Traditional benchmarks in Information Retrieval have historically segmented the task of Information Extraction to evaluate models independently~\cite{martinez2020information,cheng2021hacred,lu2022unified}. Recent endeavors, however, have led to the development of benchmarks designed for the holistic assessment of task resolution capabilities. 
Among these, HotPotQA~\cite{yang2018hotpotqa} necessitates the discovery of question-relevant information across paragraphs to aid in response formulation, FEVER~\cite{thorne2018fever,Thorne19FEVER2,Aly21Feverous} necessitates the identification of evidentiary support to validate or negate a claim, and RECLOR~\cite{yu2020reclor}, UQuAD~\cite{kazi2021uquad1}, BIOMRC~\cite{lu2022contextual} emphasizes the extraction of text segments pivotal for answering queries.
Nonetheless, the linkage between key information and queries within these benchmarks is overtly conspicuous, allowing for the location of pertinent data through string-matching techniques and facilitating correct answer derivation via one or two inferential leaps.

However, the unique feature of the DetectBench is its reliance on evidence that is widely dispersed and implicit to answer questions.

\subsection{Commonsense Reasoning}

The exploration of Commonsense Reasoning encompasses a variety of research efforts, traditionally classified into single-hop reasoning, multi-hop reasoning, and reasoning that is uncommon yet plausible.
Datasets facilitating single-hop reasoning, such as HellaSwag~\cite{zellers2019hellaswag} and WinoGrande~\cite{sakaguchi2021winogrande}, present challenges in reasoning through narrative continuation, where the difficulty often resides in the formulation of options and potentially in the design of adversarial options aimed at undermining specific models.
Multi-hop reasoning benchmarks like StrategyQA~\cite{geva2021did} annotate the reasoning path, concentrating on the capacity of models to execute multi-hop reasoning in response to questions.
Reasoning that is uncommon yet feasible, as demonstrated in datasets like α-NLG~\cite{bhagavatula2019aNLG}, d-NLI~\cite{rudinger2020bNLG}, and UnCommonsense Reasoning~\cite{zhao2023uncommonsense, arnaout2022uncommonsense}, typically originates from pre-existing datasets by selecting the least likely option as the correct response and elucidating the rationale behind this selection.

The DetectBench is categorized as uncommon but plausible multi-step reasoning, which features finding where to start such reasoning tasks.
The process of reasoning usually starts with small details that might seem unimportant.
However, when looked at more closely, these details help show a clear path that leads to a clear answer.

% \input{tables/tab_dataset_comparsion}

% The DetectBench categorizes itself as an uncommon but plausible framework for multi-hop reasoning.
% It focuses on pinpointing the starting point for multi-hop reasoning.
% Reasoning in the benchmark usually begins with apparently trivial details that, upon deeper investigation, reveal a logical path to an accurate conclusion.

\section{Detective Benchmark}
\label{sec:ben}

\begin{table}[t]
\centering
\resizebox{\columnwidth}{!}{
\begin{tabular}{>{\raggedright\arraybackslash}m{4cm}>{\centering\arraybackslash}m{2cm}>{\centering\arraybackslash}m{1.5cm}>{\centering\arraybackslash}m{1cm}}
\toprule
\multirow{2}{*}{\textbf{Dataset}} &
\textbf{Answer Correctly} &
 \multicolumn{2}{c}{\textbf{Evidence Detection}}\\
\cline{2-4}
& \textbf{Accuracy} & \textbf{Accuracy} & \textbf{RougeL} \\
\hline
DetectBench-Dev & 74.1 &63.8& 64.7 \\ \hline
DetectBench-Test-Hard & 71.8&62.4 & 65.1 \\ 
\bottomrule
\end{tabular}}
% \vspace{-2mm}
\caption{
Human performance in StructBench.}
\label{tab:human_experiments}
% \vspace{-4mm}
\end{table}

\begin{table}[t]
\centering
\resizebox{\columnwidth}{!}{
% \begin{tabular}{l{1cm}p{1.3cm}p{1.4cm}p{1cm}p}
\begin{tabular}{>{\raggedright\arraybackslash}m{2cm}>{\centering\arraybackslash}m{1.3cm}>{\centering\arraybackslash}m{1.3cm}>{\centering\arraybackslash}m{1.4cm}>{\centering\arraybackslash}m{1cm}}
\toprule
\textbf{Name} & 
\textbf{\#Sample} &
\textbf{Avg \#Token} &
\textbf{Avg \#Evidence} &
\textbf{Avg \#Jumps} \\
\hline
train&365&177&4.27&7.10\\
\hline
dev&1,770&178&4.34&7.13\\
\hline
test-normal&1,193&179&4.24&7.03\\
\hline
\hline
test-hard&300&261&7.79&13.83\\
\hline
test-distract&300&10,779&4.16&7.27\\
\hline
\hline
\textbf{All}&\textbf{3,928}&\textbf{994}&\textbf{4.55}&\textbf{7.62}\\
\bottomrule
\end{tabular}
}
% \vspace{-2mm}
\caption{
Statistic information of DetectBench.}
\label{tab:statistic}
% \vspace{-7mm}
\end{table}

\begin{figure*}[t]
    \centering
    \resizebox{\textwidth}{!}{
    \includegraphics{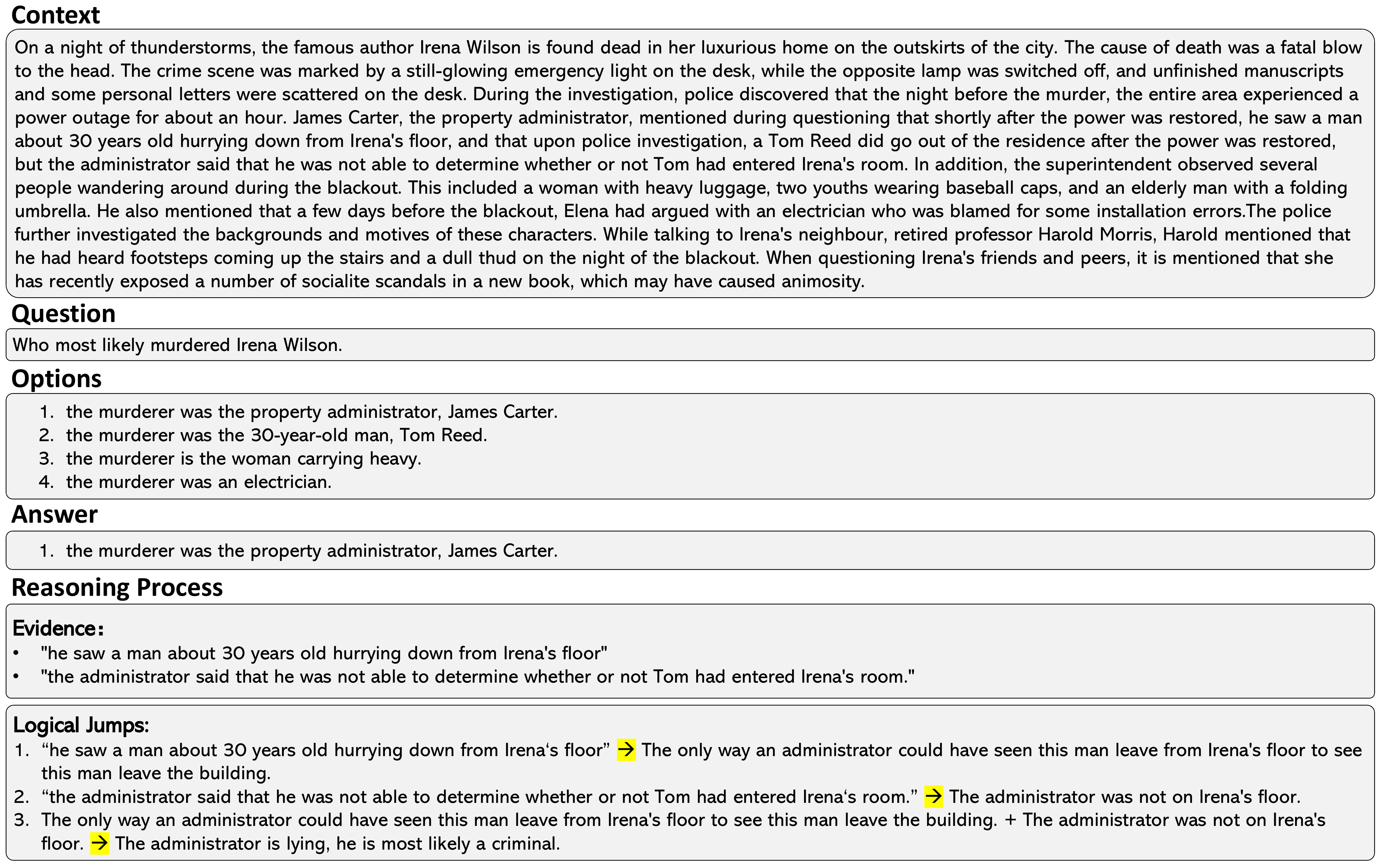}
    }
    \caption{The example of the question in DetectBench.}
    \label{fig:example}
% \vspace{-2mm}
\end{figure*}

\begin{figure*}[t]
    \centering
    \resizebox{\textwidth}{!}{
    \includegraphics{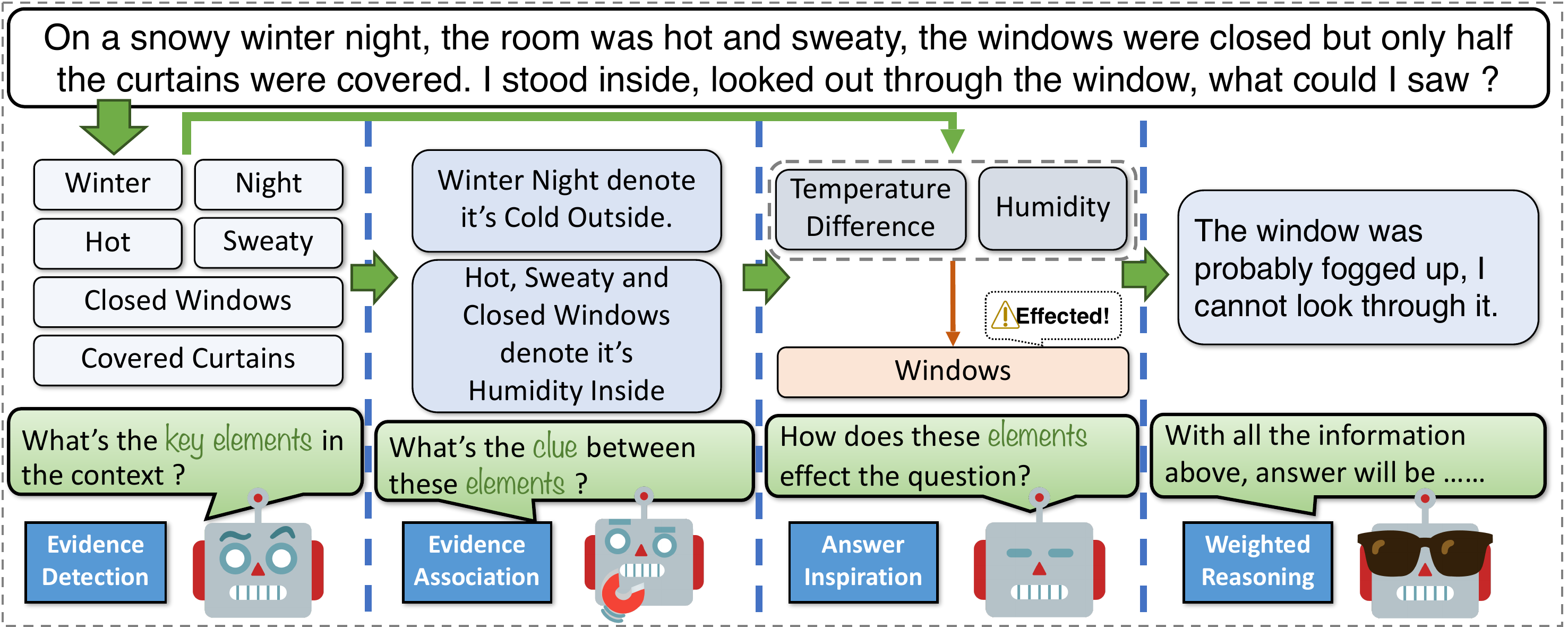}
    }
    \caption{
    The figure represents the conceptual framework of ``Detective Reasoning''.
    % The ``Detective Reasoning Prompt'' method involves providing instructions to an LLM, requiring it to output its thought process directly following the question specifications described in the figure.
    % The Detective Reasoning Finetune involves self-generating data for finetuning the model based on the thought sequence delineated in the figure.
    }
    \label{fig:framework}
% \vspace{-5mm}
\end{figure*}

\subsection{Construction}

The questions in DetectBench are sourced from open-access Detective Puzzle problems, which undergo a series of selection, rewriting, and annotation to construct into the benchmark.
DetectBench aims to evaluate the model's abilities in evidence detection and multi-step commonsense reasoning.
Therefore, the benchmark should provide the following elements:
(1). Question should not contain any ethical problem.
(2). Question descriptions should contain lengthy, complex, seemingly unrelated, and even misleading information.
(3). The solution to the question should involve multi-step reasoning based on the evidence that can be directly found in the question context.
(4). The model's response to the question needs to be capable of being assessed objectively.

\textbf{Question Selection:}
To ensure the benchmark focuses on ``evidence detection'' and ``multi-step commonsense reasoning'', we thoroughly verify all questions.
Given that detective puzzles often contain questions with multiple potential answers and varying reasoning processes, we opt for questions whose answers and reasoning processes are the most rational and unique.
Simultaneously, we excluded questions that overly rely on symbolic logic or specialized knowledge because such questions cannot be solved simply by retrieving related information or evidence but also domain knowledge and special training techniques.
Specifically, we excluded five types of questions:
1. Questions that are not ethical or have sensitive content.
2. Questions requiring visual or auditory information to answer.
3. Questions that are anti-logical, have unreasonable answers, or are overly diverse.
4. Questions requiring extensive symbolic logic or domain knowledge.
5. Questions with too obvious evidence.

\textbf{Question Rewriting:}
The original puzzle may mix the problem description with the question, sometimes even directly provide the answer, or lack relevant information for reasoning.
Therefore, we first rewrite the puzzle into \underline{``Context''} and \underline{``Question''} to distinguish between the background description and the query of the question.
Then, the original free-text puzzles are converted into a multiple-choice format. 
The converted format includes \underline{``Options''} and \underline{``Answer''} fields to represent the choices and the correct answer.
We also constructed the \underline{``Reasoning Process''} to represent the reasoning process explicitly. 
We annotated evidence within the context as \underline{``Evidence''}. 
Based on the evidence, we delineated the \underline{``Logical Jumps''}, which encompasses the multi-hop reasoning process from a single or multiple pieces of evidence to the answer.

\textbf{Manual Verification:}
All questions processed by the GPT-4-turbo-1106-preview model undergo manual verification.
Five annotators are recruited to work with the authors on verification.
This includes eliminating questions with unreasonable answers or options that require significant modification.
Additionally, detailed adjustments are performed to the options and answers to make them more reasonable.
The Appendix~\ref{appendix_annotators} provides detailed requirements and examples for annotation.

\textbf{Test-Hard \& Test-Distract:}
We introduce two more challenging datasets, DetectBench-Test-Hard and DetectBench-Test-Distract (hereafter referred to as Test-Hard and Test-Distract), with each contains 300 samples, to better distinguish the performance of powerful LLMs by longer contexts, more evidence, and deeper logic jumps.
For Test-Hard, we selected 100 samples each with the longest tokens, the most evidence, and the deepest logical jumps.
For the Test-Distract dataset, we randomly selected 300 questions, each augmented with unique descriptions including character names, attire, and environmental settings.
We further enriched these samples by appending numerous unrelated Wikipedia articles.
The unique descriptions serve as protective identifiers, ensuring that the enriched content does not obscure the questions' intrinsic semantics.

\subsection{Statistic}
The statistic information is shown in Tab.~\ref{tab:statistic}.
The split of train, dev, and all test sets aligns with the current trend of using only a small amount of data for finetuning or in-context learning and a large amount of data for evaluation~\cite{Zhou2023LIMALI}.
Each question in DetectBench is organized in JSON format, comprising five main elements: ``Context'', ``Question'', ``Options'', ``Answer'' and ``Evidence Graph'' as shown in Fig.~\ref{fig:example}.

\subsection{Human Performance}

To propose a human baseline, we invited 50 participants to answer questions from the DetectBench dev set.
The examination took three hours, and participants were allowed to leave early if they completed the task.
The participants were comprised of undergraduate and graduate students from universities across China, each remunerated at rates exceeding the local minimum hourly wage and bonuses for each correctly answered question. 

To facilitate human participation, we translated the benchmark into Chinese and used an online question-and-answer platform to collect answers and measure time spent.
Expressions in Chinese or English will not have any additional impact because DetectBench mainly involves commonsense reasoning and contains no language-specific content.
Each participant answered 15 questions from a subset of 250 questions from the DetectBench dev set, which ensured that each question was answered by three different participants.
Participants are asked to choose the option they think is correct and underline the sentence that is useful to answer the question. 
The result of the human baseline is shown in Tab.~\ref{tab:human_experiments}.

\section{Detective Reasoning}
\label{sec:rea}
\subsection{Detective Reasoning Prompt}
The Detective Reasoning Prompt is intended to help the model identify crucial information and extract precise answers through progressively deeper logical reasoning, as demonstrated in Fig.~\ref{fig:framework}.
Specially, Detective Reasoning Prompt consists of four stages:
\textbf{(1) Evidence Detection}, which aims to prompt the model to uncover all evidence, whether useful or not, within the given context.  
\textbf{(2) Evidence Association} requires the model to comprehend the inherent connections between pieces of evidence in the context and generate new related thoughts based on detected evidence.
\textbf{(3) Answer Inspiration} involves identifying the evidence necessary for solving the given question and initiating reasoning around these pieces of evidence to trigger possible answers.
\textbf{(4) Weighted Reasoning} reinforces the model's reliance on its generated reasoning process in determining the final answer compared to the overall context.
For detailed prompts for each stage, please refer to Appendix~\ref{appendix_prompt}.

\subsection{Detective Reasoning Finetune}

Building upon the aforementioned Detective Reasoning Prompt, we propose a finetuning strategy to further improve the model's evidence detection abilities.
For benchmarks that have reasoning processes explicitly annotated, such as our DetectBench, one can concatenate the reasoning outputs for each stage in the Detective Reasoning Prompt as the finetuning data. 
For benchmarks that have only standard answers, the Detective Reasoning Finetune strategy uses the other powerful LLMs to complete the reasoning process based on the questions and answers and then organize this reasoning content into the format as shown in Tab.~\ref{tab:selfquestion} in Appendix as finetuning data.

% This method has the advantage of using the freely output LLM as fine-tuning data in the first three stages. 
% This significantly reduces the complexity of constructing datasets containing inference processes.

\section{Experiments}
\label{sec:exp}

\begin{table*}[ht]
    \centering
    \resizebox{\textwidth}{!}{
    \begin{tabular}{|l|cc|cc|cc|cc|cc|cc|cc|cc|cc|}
    \hline
         &\multicolumn{2}{c|}{GPT4} &\multicolumn{2}{c|}{GPT35} &\multicolumn{2}{c|}{GLM4} &\multicolumn{2}{c|}{ChatGLM3-chat} &\multicolumn{2}{c|}{ChatGLM3-base} &\multicolumn{2}{c|}{Llama2-chat} &\multicolumn{2}{c|}{Llama2-base} \\
        &RougeL-F.&Acc.&RougeL-F.&Acc.&RougeL-F.&Acc.&RougeL-F.&Acc.&RougeL-F.&Acc.&RougeL-F.&Acc.&RougeL-F.&Acc.\\
        \hline
        
        \multicolumn{15}{|l|}{\cellcolor{mygray} \textit{Naive Questioning}} \\
        \hline
        \cellcolor{darkblue!25}Naive                           &\cellcolor{lightblue!25} 44.4 &\cellcolor{lightblue!25} 56.5 &\cellcolor{lightblue!25} 15.3 &\cellcolor{lightblue!25} 33.0 &\cellcolor{lightblue!25} 31.1 &\cellcolor{lightblue!25} 40.2 &\cellcolor{lightblue!25} 15.3 &\cellcolor{lightblue!25} 41.3 &\cellcolor{lightblue!25} 9.71 &\cellcolor{lightblue!25} 39.6 &\cellcolor{lightblue!25} 10.8 &\cellcolor{lightblue!25} 47.5 &\cellcolor{lightblue!25} 10.7 &\cellcolor{lightblue!25} 39.6 \\
        \cellcolor{darkblue!25}Naive (3-shot)                  &\cellcolor{lightblue!25} 40.6 &\cellcolor{lightblue!25} 54.4 &\cellcolor{lightblue!25} 15.3 &\cellcolor{lightblue!25} 34.9 &\cellcolor{lightblue!25} 30.3 &\cellcolor{lightblue!25} 39.4 &\cellcolor{lightblue!25} 10.8 &\cellcolor{lightblue!25} 41.8 &\cellcolor{lightblue!25} 13.1 &\cellcolor{lightblue!25} 42.3 &\cellcolor{lightblue!25} 11.5 &\cellcolor{lightblue!25} 47.1 &\cellcolor{lightblue!25} 9.9 &\cellcolor{lightblue!25} \textbf{41.4} \\

        \hline
        \multicolumn{15}{|l|}{\cellcolor{mygray} \textit{Process Enhanced Method}} \\
        \hline
        
        \cellcolor{darkblue!25}Self-CoT                        &\cellcolor{lightblue!25} 31.4 &\cellcolor{lightblue!25} 60.7 &\cellcolor{lightblue!25} 17.73 &\cellcolor{lightblue!25} 32.3 &\cellcolor{lightblue!25} 31.0 &\cellcolor{lightblue!25} 45.1 &\cellcolor{lightblue!25} 17.0 &\cellcolor{lightblue!25} 40.4 &\cellcolor{lightblue!25} 21.8 &\cellcolor{lightblue!25} 35.4 &\cellcolor{lightblue!25} 20.6 &\cellcolor{lightblue!25} 50.6 &\cellcolor{lightblue!25} 16.6 &\cellcolor{lightblue!25} 38.7 \\
        \cellcolor{darkblue!25}Auto-CoT (3-shot)               &\cellcolor{lightblue!25} 37.5 &\cellcolor{lightblue!25} 56.7 &\cellcolor{lightblue!25} 19.91 &\cellcolor{lightblue!25} 33.9 &\cellcolor{lightblue!25} \textbf{35.5} &\cellcolor{lightblue!25} 43.2 &\cellcolor{lightblue!25} 18.1 &\cellcolor{lightblue!25} 41.3 &\cellcolor{lightblue!25} 22.9 &\cellcolor{lightblue!25} 37.5 &\cellcolor{lightblue!25} 20.4 &\cellcolor{lightblue!25} 47.5 &\cellcolor{lightblue!25} 19.9 &\cellcolor{lightblue!25} 40.9 \\

        \hline
        \multicolumn{15}{|l|}{\cellcolor{mygray} \textit{Output Ensemble Method}} \\
        \hline
        
        \cellcolor{darkblue!25}Self-Consistency                &\cellcolor{lightblue!25} 31.7 &\cellcolor{lightblue!25} 54.8 &\cellcolor{lightblue!25} 18.9 &\cellcolor{lightblue!25} 33.0 &\cellcolor{lightblue!25} 25.9 &\cellcolor{lightblue!25} \textbf{49.4} &\cellcolor{lightblue!25} 14.4 &\cellcolor{lightblue!25} 40.3 &\cellcolor{lightblue!25} 25.1 &\cellcolor{lightblue!25} 37.6 &\cellcolor{lightblue!25} 19.3 &\cellcolor{lightblue!25} 41.1 &\cellcolor{lightblue!25} 25.2 &\cellcolor{lightblue!25} 39.7 \\
        \cellcolor{darkblue!25}Complexity-CoT                  &\cellcolor{lightblue!25} 28.6 &\cellcolor{lightblue!25} 61.9 &\cellcolor{lightblue!25} 20.0 &\cellcolor{lightblue!25} 34.1 &\cellcolor{lightblue!25} 28.1 &\cellcolor{lightblue!25} 44.8 &\cellcolor{lightblue!25} 17.0 &\cellcolor{lightblue!25} 40.6 &\cellcolor{lightblue!25} \textbf{23.7} &\cellcolor{lightblue!25} 34.3 &\cellcolor{lightblue!25} 21.8 &\cellcolor{lightblue!25} 50.4 &\cellcolor{lightblue!25} 29.5 &\cellcolor{lightblue!25} 40.1 \\       

        \hline
        \multicolumn{15}{|l|}{\cellcolor{mygray} \textit{Multi-step Chain-of-Thought}} \\
        \hline
        \cellcolor{darkblue!25}PS-CoT                          &\cellcolor{lightblue!25} 21.3 &\cellcolor{lightblue!25} 52.8 &\cellcolor{lightblue!25} 17.9 &\cellcolor{lightblue!25} 34.1 &\cellcolor{lightblue!25} 21.8 &\cellcolor{lightblue!25} 46.1 &\cellcolor{lightblue!25} 16.4 &\cellcolor{lightblue!25} \textbf{42.5} &\cellcolor{lightblue!25} 18.1 &\cellcolor{lightblue!25} 39.1 &\cellcolor{lightblue!25} 16.0 &\cellcolor{lightblue!25} 51.1 &\cellcolor{lightblue!25} \textbf{23.2} &\cellcolor{lightblue!25} 38.5 \\
        \cellcolor{darkblue!25}\textbf{DR Prompt~(ours)}   &\cellcolor{lightblue!25} \textbf{45.5} &\cellcolor{lightblue!25} \textbf{61.5} &\cellcolor{lightblue!25} \textbf{20.9} &\cellcolor{lightblue!25} \textbf{36.4} &\cellcolor{lightblue!25} 20.1 &\cellcolor{lightblue!25} 45.1 &\cellcolor{lightblue!25} \textbf{18.9} &\cellcolor{lightblue!25} 42.2 &\cellcolor{lightblue!25} 22.3 &\cellcolor{lightblue!25} \textbf{43.8} &\cellcolor{lightblue!25} \textbf{25.2} &\cellcolor{lightblue!25} \textbf{52.4} &\cellcolor{lightblue!25} 20.7 &\cellcolor{lightblue!25} 40.5 \\

        \hline
        \multicolumn{15}{|l|}{\cellcolor{mygray} \textit{w/ Extra Information}} \\
        \hline
        
        \cellcolor{darkyellow!25}Naive w/ Evidence               &\cellcolor{lightyellow!25} 65.4 &\cellcolor{lightyellow!25} 64.8 &\cellcolor{lightyellow!25} 42.9 &\cellcolor{lightyellow!25} 34.9 &\cellcolor{lightyellow!25} 48.3 &\cellcolor{lightyellow!25} 58.1 &\cellcolor{lightyellow!25} 22.7 &\cellcolor{lightyellow!25} 47.9 &\cellcolor{lightyellow!25} 47.1 &\cellcolor{lightyellow!25} 44.5 &\cellcolor{lightyellow!25} 48.7 &\cellcolor{lightyellow!25} 47.6 &\cellcolor{lightyellow!25} 61.3 &\cellcolor{lightyellow!25} 48.9 \\
        \cellcolor{darkyellow!25}Naive w/ Evidence (3-shot)      &\cellcolor{lightyellow!25} 63.6 &\cellcolor{lightyellow!25} 40.1 &\cellcolor{lightyellow!25} 39.5 &\cellcolor{lightyellow!25} 45.6 &\cellcolor{lightyellow!25} 43.7 &\cellcolor{lightyellow!25} 45.5 &\cellcolor{lightyellow!25} 35.8 &\cellcolor{lightyellow!25} 50.2 &\cellcolor{lightyellow!25} 31.6 &\cellcolor{lightyellow!25} 49.7 &\cellcolor{lightyellow!25} 32.5 &\cellcolor{lightyellow!25} 48.3 &\cellcolor{lightyellow!25} 67.4 &\cellcolor{lightyellow!25} 49.6 \\
        \cellcolor{darkyellow!25}Naive w/ Answer                 &\cellcolor{lightyellow!25} 47.3 &\cellcolor{lightyellow!25} 99.0 &\cellcolor{lightyellow!25} 20.3 &\cellcolor{lightyellow!25} 94.5 &\cellcolor{lightyellow!25} 36.5 &\cellcolor{lightyellow!25} 98.0 &\cellcolor{lightyellow!25} 23.0 &\cellcolor{lightyellow!25} 57.0 &\cellcolor{lightyellow!25} 18.0 &\cellcolor{lightyellow!25} 69.4 &\cellcolor{lightyellow!25} 17.9 &\cellcolor{lightyellow!25} 47.9 &\cellcolor{lightyellow!25} 13.7 &\cellcolor{lightyellow!25} 56.9 \\
        \cellcolor{darkyellow!25}Naive w/ Answer (3-shot)        &\cellcolor{lightyellow!25} 55.3 &\cellcolor{lightyellow!25} 77.6 &\cellcolor{lightyellow!25} 18.3 &\cellcolor{lightyellow!25} 82.5 &\cellcolor{lightyellow!25} 35.1 &\cellcolor{lightyellow!25} 97.0 &\cellcolor{lightyellow!25} 20.8 &\cellcolor{lightyellow!25} 49.6 &\cellcolor{lightyellow!25} 16.3 &\cellcolor{lightyellow!25} 71.3 &\cellcolor{lightyellow!25} 14.9 &\cellcolor{lightyellow!25} 35.5 &\cellcolor{lightyellow!25} 14.9 &\cellcolor{lightyellow!25} 61.1 \\
    \hline
    \end{tabular}
    }
    \caption{
    The performance of baseline models under renowned prompt methods is presented.
    Results in bold indicate the best results achieved without extra information.
    }
    \label{tab:prompt_experiments}
\vspace{-3mm}
\end{table*}

\begin{table}[t]
\centering
\resizebox{\columnwidth}{!}{
\begin{tabular}{|l|cc|cc|}
\hline
 &
 \multicolumn{2}{c|}{\textbf{Evidence Detection}} & \multicolumn{2}{c|}{\textbf{Correct Answering}} \\
 &DetectBench &HotPotQA &DetectBench &ReClor  \\
 \hline
\multicolumn{5}{|l|}{\cellcolor{mygray} \textit{ChatGLM3-Base}} \\
\hline
Naive                           &  9.7 & 26.8 & 39.6 & 30.1 \\
\textbf{DR Prompt}              & 22.3 & 25.4 & 43.8 & 31.9 \\ 
\textbf{DR FT w/ Detective}     & \textbf{37.6} & \textbf{34.2} & \textbf{50.8} & \textbf{36.7} \\
\textbf{DR FT w/ Generated}     & 35.4 & 30.9 & 43.6 & 32.9 \\  
\hline
\multicolumn{5}{|l|}{\cellcolor{mygray} \textit{ChatGLM3-Chat}} \\
\hline
Naive                           & 15.3 & 31.8 & 41.3 & 33.0 \\
\textbf{DR Prompt}              & 18.9 & 37.6 & 42.2 & 38.9 \\ 
\textbf{DR FT w/ Detective}     & \textbf{27.1} & \textbf{42.3} & \textbf{56.3} & \textbf{41.7} \\
\textbf{DR FT w/ Generated}     & 24.6 & 38.5 & 43.5 & 39.1 \\ 
\hline
\multicolumn{5}{|l|}{\cellcolor{mygray} \textit{Llama2-base}} \\
\hline
Naive                           & 10.8 & 30.6 & 47.5 & 36.7 \\
\textbf{DR Prompt}              & 20.7 & 32.1 & 40.5 & 37.5 \\ 
\textbf{DR FT w/ Detective}     & 38.6 & 37.2 & 56.7 & 39.6 \\
\textbf{DR FT w/ Generated}     & 32.4 & 32.8 & 44.6 & 33.5 \\ 
\hline
\multicolumn{5}{|l|}{\cellcolor{mygray} \textit{Llama2-Chat}} \\
\hline
Naive                           & 10.8 & 36.3 & 47.5 & 38.8 \\
\textbf{DR Prompt}              & 25.2 & 39.7 & 52.4 & 42.6 \\ 
\textbf{DR FT w/ Detective}     & \textbf{40.9} & \textbf{41.7} & \textbf{58.3} & \textbf{45.5} \\
\textbf{DR FT w/ Generated}     & 34.6 & 38.6 & 50.5 & 37.1 \\ 
\hline
\end{tabular}
}
\caption{
A detailed comparison of baseline models' performances utilizing Detective Reasoning Prompt and Fine-tuning.
% is provided.
% Outcomes in bold signify the most superior results within the same model under these experimental conditions.
}
\label{tab:DR_experiments}
\vspace{-3mm}
\end{table}

\begin{table}[t]
\centering
\resizebox{\columnwidth}{!}{
\begin{tabular}{|l|cc|cc|}
\hline
 &\multicolumn{2}{c|}{\textbf{Evidence Detection}} & \multicolumn{2}{c|}{\textbf{Correct Answering}} \\
 & Test-Hard & Test-Distract & Test-Hard & Test-Distract  \\
\hline
\multicolumn{5}{|l|}{\cellcolor{mygray} \textit{RAG-Based Framework}} \\
\hline
BM25 + GPT4                     & 16.4 & 8.8 & 27.1 & 37.0 \\
GPT4-Retriever + GPT4           & 35.4 & 19.4 & 32.2 & 37.8 \\
\hline
\multicolumn{5}{|l|}{\cellcolor{mygray} \textit{GPT4}} \\
\hline
Naive                           & 31.4 & 11.6 & 29.6 & 36.3 \\
\textbf{DR Prompt}              & \textbf{37.9} & 27.6 & \textbf{34.1} & \textbf{39.7} \\ 
\hline
\multicolumn{5}{|l|}{\cellcolor{mygray} \textit{Llama2-Chat}} \\
\hline
Naive                           & 6.5 & 3.1 & 28.3 & 26.4 \\
\textbf{DR Prompt}              & 11.7 & 15.9 & 28.7 & 30.6 \\ 
\textbf{DR FT w/ Detective}     & 17.6 & 21.8 & 35.3 & 38.5 \\
\textbf{DR FT w/ Generated}     & 15.4 & \textbf{28.6} & 31.7 & 37.1 \\ 
\hline

\end{tabular}
}
\caption{
Performance on Test-Hard and Test-Distract.
}
\label{tab:DR_hard_experiments}
\vspace{-7mm}
\end{table}

% \multicolumn{5}{|l|}{\cellcolor{mygray} \textit{ChatGLM3-Chat}} \\
% \hline
% Naive                           & - & - & - & - \\
% \textbf{DR Prompt}              & - & - & - & - \\ 
% \textbf{DR FT w/ Detective}     & - & - & - & - \\
% \textbf{DR FT w/ Generated}     & - & - & - & - \\ 
% \hline

\begin{figure}[t]
    \centering
    \resizebox{\columnwidth}{!}{
    \includegraphics{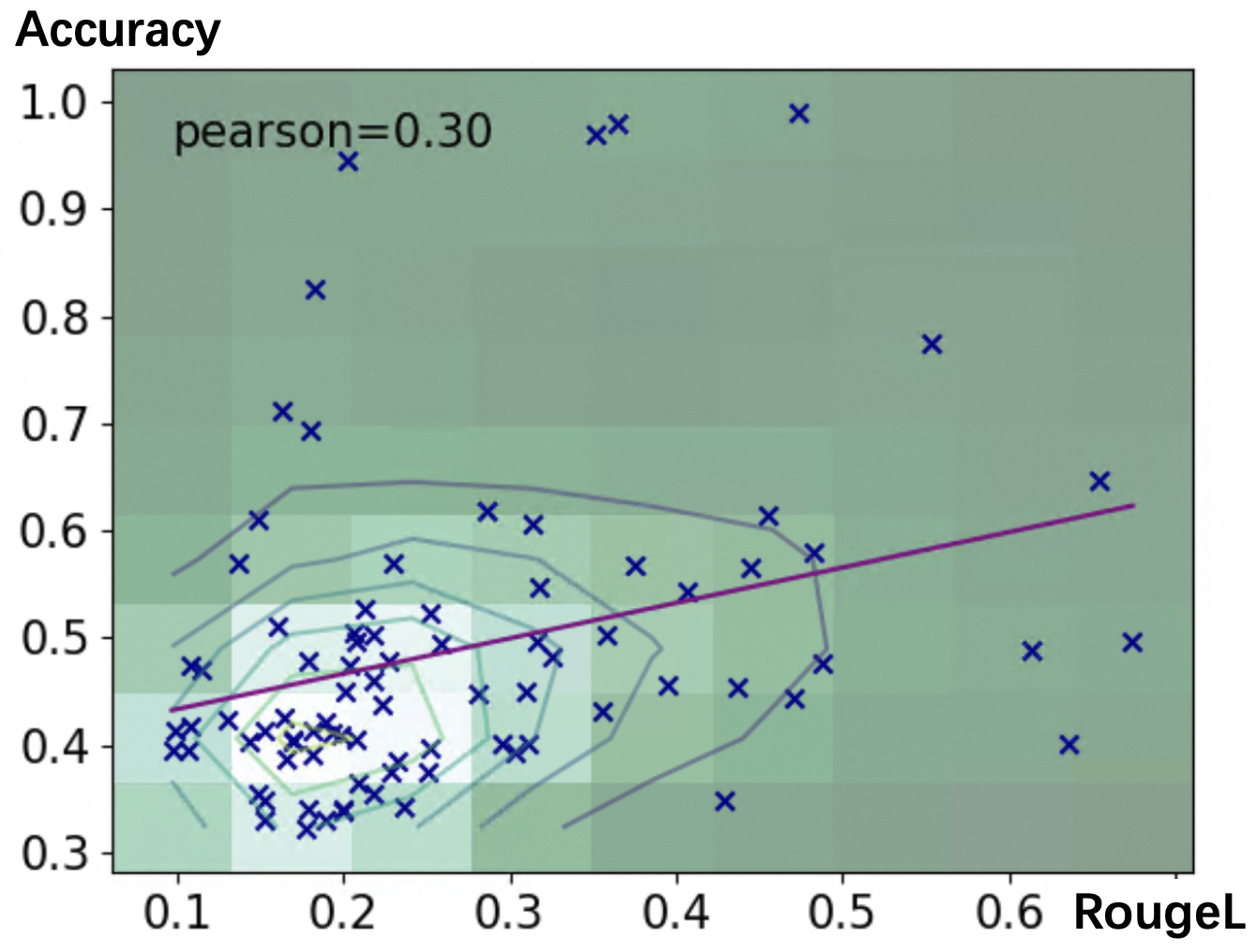}}
    \caption{The Pearson Correlation between the evidence detection~(RougeL) and reasoning performance~(Accuracy) across all models and prompt methods.}
    \label{fig:perason}
\vspace{-7mm}
\end{figure}

\subsection{Overall Setup}

\textbf{LLM Baselines:}
To test the best performance of the LLMs and ensure replicability, we have used a number of eminent models from both the API-based and the open-source domains.
These include GPT4-turbo (\textbf{GPT4})~\cite{openai2023gpt4}, GPT3.5-turbo (\textbf{GPT35})~\cite{openaichatgpt}, Llama2-7b-Base (\textbf{llama2-base}), Llama2-7b-Chat (\textbf{llama2-chat})~\cite{touvron2023llama}, GLM4 (\textbf{GLM4})~\cite{zheng2023judging}, ChatGLM3-6b-Base (\textbf{chatglm3-base}), and ChatGLM3-6B-Chat (\textbf{chatglm3-chat})~\cite{xu2023wizardlm}. The experimentation was conducted using the official APIs for GPT4-turbo, GPT-3.5-turbo, and GLM-4 between January 10 and January 29, 2024.

\textbf{Detective Reasoning:}
Our focus is on evaluating the effectiveness of the Detective Reasoning Prompt~(\textbf{DR Prompt}), fine-tuning using DetectBench data~(\textbf{DR FT w/ Detective}), and self-generated fine-tuning data based on DetectBench context, question, and answer~(\textbf{DR FT w/ Generated}). 
A subset of 398 samples from the training dataset was used for fine-tuning over three epochs with the AdamW optimizer, as detailed in Appendix~\ref{appendix_training}.
Appendix~\ref{appendix_prompt} provides detailed descriptions of the prompts used in each method.

\begin{figure*}[!ht]
    \centering
    \resizebox{\textwidth}{!}{
    \includegraphics{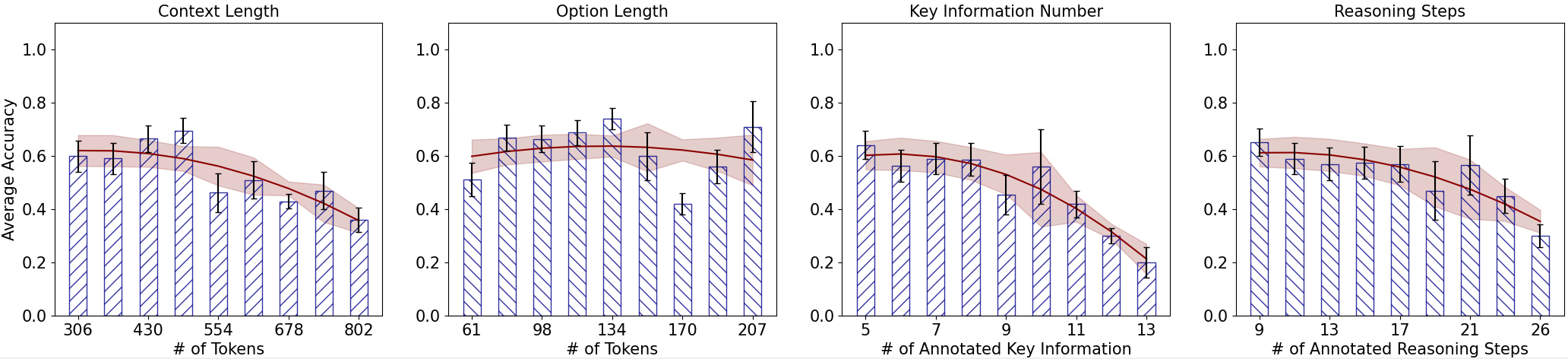}
    }
    \vspace{-6mm}
    \caption{The performance of GPT4-Turbo is correlated with the context length, option length, the number of evidence, and the number of reasoning steps involved.}
    \label{fig:gpt4_performance}
\vspace{-5mm}
\end{figure*}

\textbf{Prompt Baselines:}
% A range of prompt engineering methods were analyzed for comparative insights:
\texttt{\textbf{Naive}}, which simply inputs ``Context'', ``Question'', and ``Options'' into LLMs for answers.
\texttt{\textbf{Self-CoT}}~\cite{kojima2022large}, applying a step-by-step reasoning prompt.
\texttt{\textbf{Auto-CoT}}~\cite{zhang2022automatic}, which automates Chain of Thought (CoT) demonstrations, evaluated in a three-shot setting due to its non-zero-shot design.
\texttt{\textbf{Self-Consistency}}~\cite{wang2022self}, summarizing multiple outputs from the same model to derive a final answer.
\texttt{\textbf{Complexity-CoT}}~\cite{fu2022complexity}, selecting the longest reasoning steps among all outputs.
\texttt{\textbf{Plan-and-Solve CoT (PS-CoT)}}~\cite{wang2023plan}, focusing on problem deconstruction before solution.
\texttt{\textbf{Detective Reasoning Prompt}} is our method.
\texttt{\textbf{Naive /w Evidence}} and \texttt{\textbf{Naive /w Answer}}, enhancing inputs with ``Evidence'' and the ``Answer'' respectively.

Some methods are not included in the experiments:
Methods that involve a self-checking process, such as Tree of Thought~\cite{yao2023tot} and Graph of Thought~\cite{besta2023got}, were excluded because common sense reasoning is challenging to self-check during intermediate processes.
Methods such as Reflexion~\cite{shinn2023reflexion}, which increase the probability of a correct answer by injecting model error, were ruled out due to the prior information that would be incurred in choosing options in an option-based QA setting.

\textbf{Demonstration:}
Demonstration is about giving some examples in the context to improve LLM's understanding of output format and knowledge acquisition.
Naive Prompt appends answers after training data examples, while Auto-CoT guides the LLM in generating reasoning processes aligned with the ``Context'', ``Question'', and ``Answer''.

\textbf{Metrics:}
We evaluate the reasoning ability of LLMs based on the \textbf{Accuracy~(Acc.)} in answering the multiple-choice question on DetectBench and Reclor.
HotpotQA proposes to use F1 and Exact Match scores to evaluate models on extracting answers directly from the given context.
However, considering that the current mainstream conversational LLMs struggle to generate content identical to the original text directly, we propose to use \textbf{RougeL-F.} for evaluation on evidence detection.

\begin{table*}[t]
    \centering
    \tiny
    \resizebox{\textwidth}{!}{
\begin{tabular}{>{\raggedright\arraybackslash}m{1cm}>{\raggedright\arraybackslash}m{1cm}>{\centering\arraybackslash}m{0.5cm}>{\raggedright\arraybackslash}m{12.5cm}}

      \toprule
      \textbf{Base Model}  & \textbf{Baseline} & \textbf{Choice} & \textbf{Detected Evidence} \\
      \hline
        GPT4 & Naive & 4~\red{$\times$} & \red{$\times$}1. "Irena had argued with an electrician who was blamed for some installation errors."
\red{$\times$}2. "footsteps coming up the stairs and a dull thud on the night of the blackout."
\red{$\times$}3. "a few days before the blackout, Elena had argued with an electrician." \\
        \hline
        GPT4 & DR Prompt & 2~\red{$\times$} & 
\red{$\times$}1. "During the investigation, police discovered that the night before the murder, the entire area experienced a power outage for about an hour."
\red{$\times$}2. "James Carter, the property administrator, mentioned during questioning that shortly after the power was restored, he saw a man about 30 years old hurrying down from Irena's floor."
\red{$\times$}3. "Harold mentioned that he had heard footsteps coming up the stairs and a dull thud on the night of the blackout."
\red{$\times$}4. "When questioning Irena's friends and peers, it is mentioned that she has recently exposed a number of socialite scandals in a new book, which may have caused animosity."
 \\
        \hline
        Llama2-Chat & Naive & NaN~\red{$\times$} & \red{$\times$}1.The description given does not mention who murdered Irena Wilson. \\
        \hline
        Llama2-Chat & DR Prompt & 2~\red{$\times$} & 
        \red{$\times$}1. A man, Tom Reed, was seen hurrying down from Irena's floor after the power was restored.
        \red{$\times$}2. The property administrator, James Carter, saw someone but couldn't confirm if it was Tom Reed entering Irena's room.
        \red{$\times$}3. Several people were wandering around during the blackout, including a woman with heavy luggage, two youths, an elderly man, and Elena arguing with an electrician.
        \red{$\times$}4. Harold Morris heard footsteps and a dull thud on the night of the blackout.
        \red{$\times$}5. Irena recently exposed socialite scandals, potentially making enemies.\\
        \hline
        Llama2-Chat & DR FT w/ Detective & 1~\mydarkgreen{$\checkmark$} & 
        \red{$\times$}1. James Carter, the property administrator, mentioned during questioning that shortly after the power was restored
        \mydarkgreen{$\checkmark$}2. he saw a man about 30 years old hurrying down from Irena's floor
        \red{$\times$}3.and that upon police investigation, a Tom Reed did go out of the residence after the power was restored,
        \mydarkgreen{$\checkmark$}4. but the administrator said that he was not able to determine whether or not Tom had entered Irena's room. \\
        \hline
        Llama2-Chat & DR FT w/ Generative & 2~\red{$\times$} & 
\red{$\times$}1. "During the investigation, police discovered that the night before the murder, the entire area experienced a power outage for about an hour."
\red{$\times$}2. "While talking to Irena's neighbour, retired professor Harold Morris, Harold mentioned that he had heard footsteps coming up the stairs and a dull thud on the night of the blackout."
\red{$\times$}3. "When questioning Irena's friends and peers, it is mentioned that she has recently exposed a number of socialite scandals in a new book, which may have caused animosity." \\
        \bottomrule
        
    \end{tabular}}
    \caption{The selected response from different llms with different methods to the question in Fig~\ref{fig:example}.}
    \label{tab:error}
    \vspace{-5mm}
\end{table*}

\subsection{Performance with Different Prompt}
Tab.~\ref{tab:prompt_experiments} displays the performance of all baseline models across different prompts. 
And the conclusions are as follows:
% Based on the results in the table, we have drawn the following conclusions:

\textbf{Current LLMs struggle with Evidence Detection:}
We notice a general insufficiency in Evidence Detection, with GPT4-Turbo's average RougeL-F score only being 44.4. Open-source models like ChatGLM3 and Llama2 have even lower scores, at 9.71 and 10.7, respectively.

\textbf{There is a correlation between Evidence Detection and model reasoning performance:}
When Evidence is directly fed into LLMs, there is a significant performance improvement.
Directly informing GPT4 of the Evidence beneficial to a question enhanced its Evidence Detection by 21\%, with a 9.3\% increase in reasoning outcomes.
Moreover, giving the Answer directly to the LLM enables it to find Evidence consistent with human annotations more accurately.
Further, we analyzed the correlation between evidence detection and the final reasoning outcomes in Fig.~\ref{fig:perason}, finding a notable positive correlation.

Additionally, we discovered that telling GPT4 the answer directly could achieve an answer accuracy rate of up to 99\%, whereas informing GPT4 directly about what the Evidence is only boosts its evidence accuracy to 65.4\%, with other LLMs performing even worse. This may be due to the difficulty LLMs face in producing relevant long texts directly upon request.

\textbf{Demonstration are unstable:}
As models become increasingly adept at interpreting complex instructions, the historical utility of demonstrations in enhancing model answer parsing has diminished. Across different prompting methods and model types, a 3-shot demonstration led to unstable performance~\cite{gu2023xiezhi}.

\textbf{Detective Reasoning Prompt is superior to other baselines:}
The Detective Reasoning Prompt significantly enhanced LLMs' evidence detection and reasoning capabilities. Compared to other prompting engineering strategies, this method improved accuracy and demonstrated a broader efficacy, thereby reinforcing its value in enhancing model understanding and reasoning abilities.

\subsection{Performance of Finetuning}

The impact of Detective Reasoning Finetuning (DR FT) on various models and datasets is presented in Tab.~\ref{tab:DR_experiments}, focusing on the following aspects:

\textbf{Joint Improvements in Evidence Detection and Reasoning Performance:}
The DR FT, employing a Detective-style finetuning approach, enhances performance across all models. For instance, RougeL-F scores for the Llama2-base model improved to 38.6 on DetectBench and 37.2 on HotPotQA. Moreover, the Llama2-Chat model demonstrated a rise in reasoning accuracy to 58.3\% on DetectBench and 45.5\% on ReClor after improvements in evidence detection, signifying enhanced precision in reasoning following more accurate evidence acquisition.

\textbf{Superiority of DetectBench in Finetuning:}
Detective Reasoning Finetuning using DetectBench data significantly augments both evidence detection and reasoning capabilities in large language models (LLMs), with a notable 15.2\% increase in evidence detection accuracy and a 10.5\% improvement in overall performance. This underscores the efficacy of DetectBench in refining the information processing and reasoning abilities of models.

\subsection{In-depth Performance Analysis}

\textbf{Factors Influencing Reasoning Performance:}
Analysis of GPT4-Turbo's performance (refer to Fig.~\ref{fig:gpt4_performance}) reveals the negative impact of increased context and option lengths on accuracy, which declines from about 65\% to 35\% as context length expands from 400 to 800 words. Annotations indicate a strong correlation between the volume of evidence, depth of reasoning, and performance metrics. Specifically, an increase in the number of evidence instances and depth of reasoning correlates with a notable decrease in model accuracy, affirming the complexity-performance relationship.

\textbf{Evidence Detection Prior to Reasoning:}
Table~\ref{tab:error} showcases the differential ability of various LLMs to respond to queries.
Under the Naive Prompt, GPT-4 can identify evidence but fails in correct question answering.
Conversely, with the DR Prompt, all LLMs show deeper thinking, but still cannot find right evidence and answer.
While with DR FR w/ Detective method, Llama2-Chat, which is a 7B LLM, gets right answer and evidence.

\textbf{Impact of Increased Evidence and Complexity:}
On the Test-Hard dataset, as shown in Tab~\ref{tab:DR_hard_experiments}, models generally underperform compared to standard datasets. For instance, the GPT4-Retriever + GPT4 combination leads in evidence detection at 35.4\% accuracy, yet falls short of its developmental benchmarks. The Test-Hard dataset, characterized by a higher average of evidences (7.79) and jumps (13.83), significantly complicates the information synthesis for models, as evidenced by lower performance metrics, including a 17.6\% accuracy in evidence detection and a 35.3\% correct answering rate under the DR FT w/ Detective setting.

\textbf{Challenges Posed by Long Contexts:}
In the Test-Distract dataset, as shown in Tab~\ref{tab:DR_hard_experiments}, long texts pose considerable challenges, exemplified by a plummet in BM25 + GPT4's evidence detection accuracy to 8.8\%, a stark contrast to its performance on the Test-Hard dataset. The average token count in this dataset is 10,779, overwhelming model capacities and lowering accuracy. However, models like Llama2-Chat, enhanced through DR FT w/ Generated method, show a better evidence detection accuracy of 28.6\% on this dataset, indicating that specialized finetuning can partly mitigate the effects of long texts, though not completely.

\section{Conclusion}
\label{sec:con}

This paper introduces the DetectBench to assess LLMs' abilities in evidence and multi-step commonsense reasoning within a long context.
We also propose a novel type of prompt and fine-tuning method named Detective Reasoning to augment LLM's performance in evidence detection and thereby augment performance in commonsense reasoning.
The experiment results show that the abilities of evidence detection and reasoning performance are correlated.
Detective Reasoning effectively enhances the capability of LLMs in evidence detection, thereby improving the LLMs' commonsense reasoning results in long text contexts.

% \newpage
\section{Limitations}

DetectBench is designed to facilitate LLMs' abilities in Evidence Detection and Multi-hop Commonsense Reasoning within long contexts. 
However, compared to the information in real-world scenarios, the complexity and breadth of data in DetectBench are noticeably insufficient. 
Implementing Detective Reasoning has been proven to effectively enhance the Evidence Detection capability of LLMs, thereby improving reasoning performance. 
However, this strategy is primarily suitable for tasks that require extracting and reasoning about relevant Evidence from long contexts. 
If applied in short-text scenarios, where it is necessary to combine implicit knowledge gained from common sense or experiential understanding, its effectiveness would be significantly reduced.

\section{Ethical Concerns}

Considering that Detective Puzzles may contain many sensitive topics, including but not limited to murder, theft, deception, etc.
Existing LLMs might refuse to answer sensitive questions for safety reasons, putting those LLMs that prioritize higher safety standards at a disadvantage when assessed using Detective Puzzles.
Additionally, fine-tuning LLMs on such data could inadvertently amplify security vulnerabilities.

To mitigate ethical dilemmas associated with detective reasoning benchmarks, we have invested significant effort and resources to achieve a dual objective: ensuring that models committed to safety do not refuse to answer sensitive questions; and ensuring that the use of DetectBench does not compromise the safety of the models.

\bibliography{anthology,custom}
\appendix
\section{Training Details}
\label{appendix_training}
For the models llama2-7b-base, llama2-7b-chat, ChatGPT3-6b-base, and ChatGPT3-6b-chat, we executed two distinct training methodologies:
\begin{enumerate}
    \item Directly utilizing the training data from the Detective Reasoning Benchmark to compose the Detective Reasoning Finetune data.
    \item Employing the ``Context'', ``Question'', and ``Answer'' in Detective Reasoning Benchmark to automatically generate Detective Reasoning Finetune data.
\end{enumerate}

The specific training parameters are detailed in Tab.~\ref{tab:training_param}.

\begin{table*}[!ht]
    \centering
    \myred{
    \begin{tabular}{|c|c|c|c|c|}
    \hline
    \multicolumn{5}{|c|}{Training Detail} \\
    \hline
    \# of Samples & \# of Tokens & \# of epochs & warm\_up steps & learning rate \\
    396 & 162,868 & 3 & 200 & 1e-5  \\
    \hline
    \end{tabular}
    }
    \caption{\myred{All the parameter setting in the training process.}}
    \label{tab:training_param}
\end{table*}

\begin{table*}[!ht]
\centering
\resizebox{\textwidth}{!}{
\begin{tabular}{lrrccc}
\hline
Benchmark & \# of Questions & \# of Context & Explanation to Answer & Ansering Format & Metrics \\ \hline

HotpotQA~\cite{yang2018hotpotqa}    & 90,564   & 873 & & Free Text & Rouge     \\ \hline
% HotpotQA-Distractor & 7,405  & 3,259 & & Free Text & Rouge     \\ \hline
% FEVER~\cite{thorne2018fever}    &  185,445 & 9.4 & \mydarkgreen{$\checkmark$} &\makecell[l]{Classification\\\&Text retrieval} & Accuracy \& F1 \\ \hline

HellaSwag~\cite{zellers2019hellaswag}   & 59,950    & 38  & & Choice QA & Accuracy  \\ \hline

Reclor~\cite{yu2020reclor}      & 6,138     & 66  & & Choice QA & Accuracy     \\ \hline

WinoGrande~\cite{sakaguchi2021winogrande}  & 12,282    & 21  & \mydarkgreen{$\checkmark$}  & Choice QA & Accuracy  \\ \hline

WikiNLDB-25~\cite{thorne2021database} & 5,252 & 221 & & Free Text & EM \& Rouge\\ \hline

WikiNLDB-1000 & 300 & 3,876 & & Free Text & EM \& Rouge\\ \hline

% FEVEROUS~\cite{Aly21Feverous} &87,026&25.3&\mydarkgreen{$\checkmark$}&\makecell[l]{Classification\\\&Text retrieval}& Accuracy \& F1\\ \hline

TopiocQA~\cite{adlakha2022topiocqa} &50,466&145&&Free Text&EM \& F1\\ \hline

LongBench~\cite{bai2023longbench} &  4,750 & 6,711 & & Free Text & F1 \& Rouge \& ED\\ \hline

\multirow{2}{*}{DetectBench} &
\multirow{2}{*}{3,928} & \multirow{2}{*}{994} & \multirow{2}{*}{\mydarkgreen{$\checkmark$}}  & Choice QA \& Free &Accuracy \\
&&&&Text Reasoning& \& Rouge\\ \hline

\multirow{2}{*}{DetectBench-Test-Distract} &
\multirow{2}{*}{300} & \multirow{2}{*}{10,779} & \multirow{2}{*}{\mydarkgreen{$\checkmark$}}  & Choice QA \& Free &Accuracy \\
&&&&Text Reasoning& \& Rouge\\ \hline

\end{tabular}}
\caption{The comparison between the DetectBench with other and Information Retrieval Benchmarks and Common Sense Reasoning Benchmarks.}
\label{tab:comparsion_statistic}
\vspace{-5mm}
\end{table*}

% BIOMRC~\cite{pappas2020biomrc} &812,707&254.0&&Choice QA &Accuracy\\ \hline
% TrivalQA~\cite{joshi2017triviaqa} &&&&&\\ \hline
% QuAC~\cite{choi2018quac} &&&&&\\ \hline
% StrategyQA~\cite{geva2021did}  & 2,780     & 9.6   & \mydarkgreen{$\checkmark$}  & Bool QA   & Accuracy  \\ \hline
 % \multirow{2}{*}{True Detective} &  \multirow{2}{*}{191} &  \multirow{2}{*}{1204} &  \multirow{2}{*}{2000+} &  \multirow{2}{*}{UNK} &  \multirow{2}{*}{UNK} &  \multirow{2}{*}{False} & Choice QA \& Free &Accuracy \\
% &&&&&&&Text Reasoning& \& Rouge\\ \hline

\section{Detail about Manual Annotation}
\label{appendix_annotators}
\subsection{Details about Annotators}
The annotators for this research are the authors of this paper themselves, who are experts in the field of Computer Science and Cognitive Psychology.
The entire annotation process was under the stringent supervision and scrutiny of the first author of this paper.
\subsection{Annotation Tasks and Goals}
The purpose of the manual annotation tasks was twofold.
The first goal was to obtain comprehensive annotated datasets that encapsulate the essential features of the target text, which could be further leveraged for tasks such as training, testing, and model evaluation.
The second goal was to provide a detailed, rigorous, and systematic assessment of the annotated data quality to assess its fit and reliability for the subsequent analysis.
All the detailed annotation tasks and targets are listed in Tab.~\ref{tab:annotation_request}.
\subsection{Case of Annotation}
In our efforts to delineate the complex annotation process and ensure the replicable rigor of experiments, this section provides an in-depth display of the manual annotation cases.
The aim is to elucidate the categorical distinctions and precise definitions adopted in the annotations, thereby facilitating fellow researchers in ascertaining the veracity of the annotated data.
Representative cases from the annotation process have been cataloged in Tab.~\ref{tab:appendix_annotation_example} for comprehensive reference and understanding.

\begin{table*}[t]
    \centering
    \resizebox{\textwidth}{!}{
    \begin{tabular}{|p{0.3\textwidth}|p{0.69\textwidth}|}
    \hline
         \textbf{Task} & \textbf{Requirements}\\
         \hline
         \multirow{5}{*}{Question Verification}
         & 1.1 Delete if answering the question requires non-text information, like audio or image. \\
         & 1.2 Delete if there is a substantial amount of mathematical content or involve of too much domain knowledge. \\
         & 1.3 Delete if there is no ample presence of daily scenarios. \\
         & 1.4 Delete if the answer is not correct. \\
         & 1.5 Delete if there is any discrimination or bias concerning gender, race, nation, or religion. \\

         \hline
         \multirow{3}{*}{Question Rewrite} 
         &2.1 Standardize the Expression. \\
         &2.2 Rewrite a decent answer to the question.\\
         &2.3 Separate ``Question''and ``Context''.\\
         &2.4 Write decent and confusing ``Options'' of the question.\\
         
         \hline
         \multirow{3}{*}{Clue Graph Construction} 
         &3.1 Regenerate or rewrite if the ``Key Information of Context'' cannot exact match to the text in ``Context''.\\
         &3.2 Regenerate or rewrite if the connection or reasoning is redundant.\\
         &3.3 Delete the question or rewrite it there lack of important reasoning processes or connections in Clue Graph.\\
    \hline
    \end{tabular}}
    \caption{All tasks that require manual annotation, along with the specific requirements for each task.}
    \label{tab:annotation_request}
\end{table*}
% \input{tables/tab_questions}
% \usepackage{multirow}
% \usepackage[normalem]{ulem}
\begin{table*}[t]
\centering
\resizebox{\textwidth}{!}{
\begin{tabular}{|c|l|l|}
\hline
Task &
  Requirements &
  Cases \\ \hline
\multirow{5}{*}{\begin{tabular}[c]{@{}c@{}}Question \\ Verification\end{tabular}}
&
\begin{tabular}[c]{@{}l@{}}Delete if answering the question \\ requires non-text information, like \\ audio or image.\end{tabular} 
&
\begin{tabular}[c]{@{}l@{}}
Context: ``Listen to the following music clip...''\\
Question: ``What instrument is playing?''\\ 
Hint: ``Consider the type of information required to answer the question.''\\ 
Answer: ``Piano''
\end{tabular} \\ \cline{2-3} 
 &
  \begin{tabular}[c]{@{}l@{}}Delete if there is a substantial \\ amount of mathematical content.\end{tabular} &
  \begin{tabular}[c]{@{}l@{}}Context: ``Consider the mathematical proof of Fermat's Last Theorem...''\\ Question: ``Can you explain the proof?''\\ Hint: ``Focus on the subject matter of the proof.''\\ Answer: ``It's a complex proof involving modular forms...''\end{tabular} \\ \cline{2-3} 
 &
  \begin{tabular}[c]{@{}l@{}}Delete if there is no ample presence \\ of daily scenarios.\end{tabular} &
  \begin{tabular}[c]{@{}l@{}}Context: ``In a quantum physics experiment...''\\ Question: ``What is the result?''\\ Hint: ``Consider the context of the experiment.''\\ Answer: ``A specific quantum state''\end{tabular} \\ \cline{2-3} 
 &
  Delete if the answer is not correct. &
  \begin{tabular}[c]{@{}l@{}}Context: ``The cat is on the roof''\\ Question: ``Where is the cat?''\\ Hint: ``Check the location mentioned in the context.''\\ Answer: ``In the garden''\end{tabular} \\ \cline{2-3} 
 &
  \begin{tabular}[c]{@{}l@{}}Delete if there is any discrimination \\ or bias concerning gender, race, \\ nation, or religion.\end{tabular} &
  \begin{tabular}[c]{@{}l@{}}Context: ``All people from X are lazy...''\\ Question: ``What are people from X like?''\\ Hint: ``Considering the description of X.''\\ Answer: ``Lazy''\end{tabular} \\ \hline
\multirow{4}{*}{\begin{tabular}[c]{@{}c@{}}Question \\ Rewrite\end{tabular}} &
  Standardize the Expression. &
  \begin{tabular}[c]{@{}l@{}}Original: ``$\langle$ /span $\rangle$ A family decides to move into the city and looks for a house. \textbackslash n \textbackslash n There are three ...'' \\
  Rewritten: ``A family decides to move into the city and looks for a house. There are three ... ''\end{tabular} \\ \cline{2-3} 
 &
  \begin{tabular}[c]{@{}l@{}}Rewrite a decent answer to the \\ question.\end{tabular} &
  \begin{tabular}[c]{@{}l@{}}Original Answer: ``This is a famous question, in my thought, the answer is ......''\\ Rewritten Answer: ``The answer is ......''\end{tabular} \\ \cline{2-3} 
 &
  Separate ``Question'' and ``Context''. &
  \begin{tabular}[c]{@{}l@{}}Original: \\  Context and Question: ``In 1862, during the American Civil War, the Battle \\         \quad                           of Antietam took place near Sharpsburg, Maryland... \\        \quad                              What was the significance of the Battle of Antietam?''\\ Separated:\\   Context: ``In 1862, during the American Civil War, the Battle of Antietam \\      \quad                      took place near Sharpsburg, Maryland...''\\   Question: ``What was the significance of the Battle of Antietam?''\end{tabular} \\ \cline{2-3} 
&
\begin{tabular}[c]{@{}l@{}}Write decent and confusing ``Options''\\ of the question.\end{tabular} &
\begin{tabular}[c]{@{}l@{}}
     Context: \\
     As the investigation unfolded, the police tape crisscrossed the snow-laden streets, casting eerie shadows under\\ \quad
     the moonlit night. The neighborhood, usually quiet and reclusive... \\
     Question: \\
     Do you think this young man is guilty or not? \\
     Answer: \\
     The young man could not have seen the murderer's detailed features due to the room's conditions \\
     Options: \\
     A) The young man was telling the truth, and the blond boyfriend was the murderer.\\
     B) The young man lied about the time of witnessing the murder to mislead the investigation.\\
     C) The young man could not have seen the murderer's detailed features due to the room's conditions.\\
     D) The victim had another visitor that night who was the real murderer\\
\end{tabular}\\ \hline

\multirow{3}{*}{\begin{tabular}[c]{@{}c@{}}Clue Graph\\Construction\end{tabular}} &
  \begin{tabular}[c]{@{}l@{}}Regenerate or rewrite if the ``Key\\ Information of Context'' cannot exact\\ match to the text in ``Context''.\end{tabular} &
  \begin{tabular}[c]{@{}l@{}}Original \\   Context: ``On a snowy winter night ...''\\   Key Information: ``On a blustery snowy winter night''\\ Rewritten\\   Key Information: ``On a snowy winter night ...''\end{tabular} \\ \cline{2-3} 
 &
  \begin{tabular}[c]{@{}l@{}}Regenerate or rewrite if the connection\\ or reasoning is redundant\end{tabular} &
  \begin{tabular}[c]{@{}l@{}}Original\\   Reasoning Process: ``Serene snowy setting + Murder at 68 King's West Road around 8pm \\ \quad→ Peaceful night disrupted by murder \\   Rewritten: \\Reasoning Process: \sout{``Serene snowy setting + Murder at 68 King's West Road around 8pm }\\ \quad \sout{→ Peaceful night disrupted by murder} \\\end{tabular} \\ \cline{2-3} 
 &
  \begin{tabular}[c]{@{}l@{}}Delete the question or rewrite it there\\ lack of important reasoning processes\\ or connections in Clue Graph.\end{tabular} &
  \begin{tabular}[c]{@{}l@{}}-\end{tabular} \\ \hline
  
% \multirow{2}{*}{\begin{tabular}[c]{@{}c@{}}GPT-4 \\ Output \\ Verification\end{tabular}} &
%   \begin{tabular}[c]{@{}l@{}}Round the length of answer divided \\ by 25 to the nearest integer, and \\ manually verify five samples from \\ each interger.\end{tabular} &
%   / \\ \cline{2-3} 
%  &
%   \begin{tabular}[c]{@{}l@{}}Evaluate whether the answer result \\ is consistent with the standard \\ answer.\end{tabular} &
%   \begin{tabular}[c]{@{}l@{}}Standard Answer: ``It's raining''\\ GPT-4 Output: ``It's sunny''\\ Evaluation: Inconsistent\end{tabular} \\ \hline
\end{tabular}}
\caption{The examples in our annotation process}
\label{tab:appendix_annotation_example}
\end{table*}

 % \hline
 %         \multirow{3}{*}{Question Rewrite} 
 %         &2.1 Standardize the Expression. \\
 %         &2.2 Separate ``Question''and ``Context''.\\
 %         &2.3 Summarize the answer to the question.\\
 %         &2.4 Write decent and confusing ``Options'' of the question.\\
         
 %         \hline
 %         \multirow{3}{*}{Clue Graph Construction} 
 %         &3.1 Regenerate or rewrite if the ``Key Information of Context'' cannot exact match to the text in ``Context''.\\
 %         &3.2 Regenerate or rewrite if the connection or reasoning is redundant.\\
 %         &3.3 Delete the question or rewrite it there lack of important reasoning processes or connections in Clue Graph.\\

\begin{table}[t]
\centering
\resizebox{\columnwidth}{!}{
\begin{tabular}{llll}
\toprule
Type                        & Example                                       &\#                       &\% \\\hline
\multirow{3}{*}{How}       &\textit{``How was the murder weapon}          &\multirow{3}{*}{1,647}  &\multirow{3}{*}{41.9}\\
                            &\textit{handled such that it was not}&&\\
                            &\textit{discovered at the scene?''}&&\\
\hline
\multirow{2}{*}{What}       &\textit{``What's the house number}             &\multirow{2}{*}{731}    &\multirow{2}{*}{18.6}\\
                            &\textit{where Smith lives?''}&&\\
\hline
\multirow{3}{*}{Which}     &\textit{``Which building doesn't have}    &\multirow{3}{*}{498}    &\multirow{3}{*}{12.7}\\
                            &\textit{any graduatestudents living in}&&\\
                            &\textit{this dormitory building?''}&&\\
\hline
\multirow{2}{*}{Who}        &\textit{``Who is the murderer of the}              &\multirow{2}{*}{459}    &\multirow{2}{*}{11.7}\\
                            &\textit{painter?''}&&\\
\hline
\multirow{1}{*}{Why}        &\textit{``Why did Harry suspect Filch?''}      &\multirow{1}{*}{378}    &\multirow{1}{*}{9.6}\\
                            % &&&\\
\hline
\multirow{1}{*}{When}       &\textit{``When is Teacher's birthday?''}       &\multirow{1}{*}{167}    &\multirow{1}{*}{4.3}\\
                            % &&&\\
\hline
\multirow{2}{*}{Where}      &\textit{``Where exactly does woman}&\multirow{2}{*}{121}   &\multirow{2}{*}{3.1}\\
                            &\textit{come from?''}&&\\
\hline
\multirow{3}{*}{Other}     &\textit{``Please determine the respective}    &\multirow{3}{*}{378}    &\multirow{3}{*}{9.6}\\
                            &\textit{professions of Faulkner, Santiago,}&&\\
                            &\textit{and Hemingway.''}&&\\
\hline
\hline
\textbf{All} && \textbf{3928} & \textbf{100} \\

\bottomrule
\end{tabular}}
\caption{
All eight types of questions in DetectBench and their frequency. 
Note that each question in DetectBench may contain different types of questions.}
\label{tab:category}
\end{table}

\section{Types of Questions in StructBench}
Tab.~\ref{tab:category} reveals a distinct preference for process-oriented questions for ``How'' to form the largest category.
Comparatively, descriptive and person-focused questions, such as ``What'', ``Which'', and ``Who'', are also notably present.

\section{Performance on Different Question Types}
\begin{figure}[!ht]
    \centering
    \resizebox{\columnwidth}{!}{
    \includegraphics{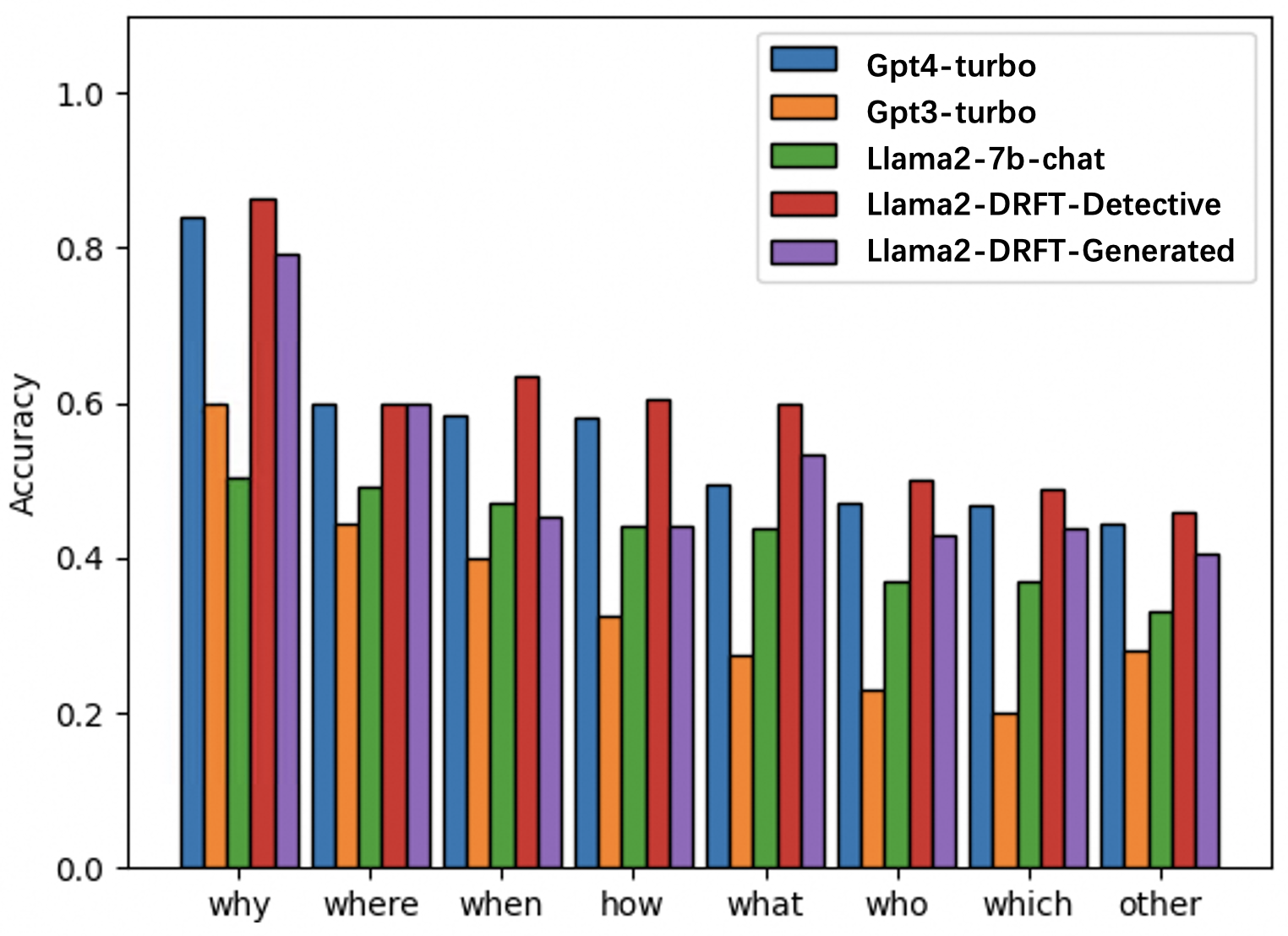}
    }
    \vspace{-6mm}
    \caption{The performance of various models varies across different Question Types.}
    \label{fig:performance_category}
% \vspace{-5mm}
\end{figure}
As shown in Fig.~\ref{fig:performance_category}, the performance differences across various question types indicate that the existing LLMs excel in answering ``why'' and ``where'' questions, with the fine-tuned Llama-2 model achieving an impressive accuracy rate of 90\%.
In contrast, the accuracy rates for ``who'', ``which'', and other types of questions hover around 50\%.
This discrepancy suggests that while the model effectively handles questions requiring an understanding of processes and environments, it struggles with questions that require complex entity recognition and relationship discernment, pointing toward directions for future model improvements.

\section{Experiments Details}
\label{appendix_experiment}

\subsection{Parameters in Inference}
Our experiments involved two types of hyperparameters.
The first type pertains to the seeds of random numbers used in various Python libraries, while the second type refers to the hyperparameters used when invoking the AutoCausalLM class from the transformers library for generation.
We configured our settings as demonstrated in Table~\ref{tab:hyperparameter}.

\begin{table*}[!ht]
    \centering
    \resizebox{\textwidth}{!}{
    \myred{
    \begin{tabular}{|c|c|c|c|c|}
    \hline
    \multicolumn{5}{|c|}{Random Seed} \\
    \hline
    torch.manual\_seed & torch.cuda.manual\_seed\_all & numpy.random.seed & random.seed & torch.backends.cudnn.deterministirc \\
    42 & 42 & 42 & 42 & True \\
    \hline
    \hline
    \multicolumn{5}{|c|}{AutoCausalLM} \\
    \hline
    temperature & top\_p & top\_k & num\_beams & max\_new\_token \\
    0.95 & 0.95 & 5 & 2 & 2000  \\
    \hline
    \end{tabular}
    }}
    \caption{\myred{All the parameter setting in model inference in our experiments.}}
    \label{tab:hyperparameter}
\end{table*}

\subsection{Prompt Details}
\label{appendix_prompt}
This section primarily showcases the prompts employed by all Prompt Engineers throughout the experiment.

Table~\ref{tab:naive} displays the \texttt{Naive} prompts,
Table~\ref{tab:naivekeyinfo} presents the \texttt{Naive w/ Key Info} prompts,
Table~\ref{tab:naiveanswer} outlines the \texttt{Naive w/ Answer} prompts,
Table~\ref{tab:selfcot} features the \texttt{Self-CoT} prompts,
Table~\ref{tab:selfconsistency} exhibits the \texttt{Self-Consistency} prompts,
Table~\ref{tab:complexitycot} reveals the \texttt{Complexity-CoT} prompts,
Table~\ref{tab:pscot} shows the \texttt{PS-CoT} prompts,
Table~\ref{tab:selfquestion} displays the \texttt{Detective Reasoning Prompt} prompts, and
% Table~\ref{tab:selfquestion_mr} showcases the \texttt{Detective Reasoning w/ Multi Round Chat} prompts.

% \section{Examples of Demonstration}
% \label{appendix_demonstration}

% Herein, we present an example of a 3-shot demonstration for each of the following approaches: \texttt{Naive (3-shot)}, \texttt{Naive w/ Key Info (3-shot)}, \texttt{Naive w/ Answer (3-shot)}, and \texttt{Auto-Cot (3-shot)}.

% Table~\ref{tab:demo_naive} illustrates an example for \texttt{Naive (3-shot)}.
% Table~\ref{tab:demo_autocot} demonstrates an example for \texttt{Auto-Cot (3-shot)}.
% Table~\ref{tab:demo_naive_keyinfo} provides an example for \texttt{Naive w/ Key Info (3-shot)}.
% Table~\ref{tab:demo_naive_answer} offers an example for \texttt{Naive w/ Answer (3-shot)}.

\begin{table*}[!ht]
    \centering
    \resizebox{\textwidth}{!}{
    \begin{tabular}{|p{\textwidth}|}
    \hline
\# -*- coding: utf-8 -*-\\

Variables:\\
!<INPUT 0>! -- Context\\
!<INPUT 1>! -- Question\\
!<INPUT 2>! -- Options\\

<commentblockmarker>\#\#\#</commentblockmarker>

Below I will give you a detective reasoning question, please summarize the key clues in this question based on the Context, the options and choose the answer you think is correct.
Note: When generating the answer, please only output the serial number of the option.

\#\#\# Context:\\
!<INPUT 0>!

\#\#\# Question:\\
!<INPUT 1>!

\#\#\# Options:\\
!<INPUT 2>!

Your output will contain the following:
\#\#\# Evidence: Please output what you consider to be the Evidence in the Context. Please note that the Evidence needs to be directly from the Context, i.e. it is a string originally in the Context that can be matched directly to the original text by string matching.
\#\#\# Answer: please output only the serial numbers.

Please follow the format below for your output:

\#\#\# Evidence:
xxxxx\\

\#\#\# Answer:
1/2/3/4\\
\bottomrule
\end{tabular}}
\caption{Prompt of \texttt{Naive} method}
\label{tab:naive}
\end{table*}

\begin{table*}[!ht]
    \centering
    \resizebox{\textwidth}{!}{
    \begin{tabular}{|p{\textwidth}|}
    \hline
\# -*- coding: utf-8 -*-\\

Variables:\\
!<INPUT 0>! -- Context\\
!<INPUT 1>! -- Question\\
!<INPUT 2>! -- Evidence
!<INPUT 3>! -- Options\\

<commentblockmarker>\#\#\#</commentblockmarker>

Below I will give you a detective reasoning question, please summarize the key clues in the question based on the Context, the options, and the answer, and choose the answer you think is correct.
Note: When generating the answer, please output only the serial number of the option.\\
\\
\#\#\# Context:\\
!<INPUT 0>!\\
\\
\#\#\# Question:\\
!<INPUT 1>!\\
\\
\#\#\# Evidence:\\
!<INPUT 2>!\\
\\
\#\#\# Option:\\
!<INPUT 3>!
\\
Your output will contain the following:\\
\#\#\# Evidence: Please output what you consider to be the Evidence in the Context. Please note that the Evidence needs to be directly from the Context, i.e. it is a string originally in the Context that can be matched directly to the original text by string matching.\\
\#\#\# Answer: please output only the serial numbers.\\
\\
Please follow the format below for your output:\\
\\
\#\#\# Evidence:\\
xxxxx\\
\\
\#\#\# Answer:\\
1/2/3/4\\
    \hline
    \end{tabular}}
    \caption{Prompt of \texttt{Naive w/ Evidence} method}
    \label{tab:naivekeyinfo}
\end{table*}

\begin{table*}[!ht]
    \centering
    \resizebox{\textwidth}{!}{
    \begin{tabular}{|p{\textwidth}|}
    \hline
\# -*- coding: utf-8 -*-\\

Variables:\\
!<INPUT 0>! -- Context\\
!<INPUT 1>! -- Question\\
!<INPUT 2>! -- Options\\
!<INPUT 3>! -- Answer\\
\\
<commentblockmarker>\#\#\#</commentblockmarker>\\
\\
Below I will give you a detective reasoning question, please summarize the key clues in the question based on the Context, the options, and the answer, and choose the answer you think is correct.\\
Note: When generating the answer, please output only the serial number of the option.\\
\\
\#\#\# Context:\\
!<INPUT 0>!\\
\\
\#\#\# Question:\\
!<INPUT 1>!\\
\\
\#\#\# Options:\\
!<INPUT 2>!\\
\\
\#\#\# Answer:
!<INPUT 3>!\\
\\
Your output will contain the following:\\
\#\#\# Evidence: Please output what you consider to be the Evidence in the Context. Please note that the Evidence needs to be directly from the Context, i.e. it is a string originally in the Context that can be matched directly to the original text by string matching.\\
\#\#\# Answer: please output only the serial numbers.\\
\\
Please follow the format below for your output:\\
\\
\#\#\# Evidence:
xxxxx\\

\#\#\# Answer:\\
1/2/3/4\\
    \hline
    \end{tabular}}
    \caption{Prompt of \texttt{Naive w/ Answer} method}
    \label{tab:naiveanswer}
\end{table*}

\begin{table*}[!ht]
    \centering
    \resizebox{\textwidth}{!}{
    \begin{tabular}{|p{\textwidth}|}
    \hline
\# -*- coding: utf-8 -*-\\
\\
Variables:\\
!<INPUT 0>! -- Context\\
!<INPUT 1>! -- Question\\
!<INPUT 2>! -- Options\\
\\
<commentblockmarker>\#\#\#</commentblockmarker>\\
\\
Below I will give you a detective reasoning question, please generate your thought process step by step based on the Context and the options and choose the answer you think is correct.\\
Note: When generating the answer, please output only the serial number of the option.\\
\\
\#\#\# Context:\\
!<INPUT 0>!\\
\\
\#\#\# Question:\\
!<INPUT 1>!\\
\\
\#\#\# Options:\\
!<INPUT 2>!\\
\\
Your output will contain the following:\\
\#\#\# Thought: please output your thinking process step by step.\\
\#\#\# Evidence: Please output what you think is the Evidence in the Context. Please note that the Evidence needs to be directly from the Context, i.e. it is a string originally in the Context that can be matched directly to the original text by string matching.\\
\#\#\# Answer: please output only the serial numbers.\\
\\
Please have your output follow the format below:\\
\\
\#\#\# Thought:\\
xxxxxx\\
\\
\#\#\# Evidence:\\
xxxxx\\
\\
\#\#\# Answers:\\
1/2/3/4\\
    \hline
    \end{tabular}}
    \caption{Prompt of \texttt{Self-CoT} method}
    \label{tab:selfcot}
\end{table*}

\begin{table*}[!ht]
    \centering
    \resizebox{\textwidth}{!}{
    \begin{tabular}{|p{\textwidth}|}
    \hline
\# -*- coding: utf-8 -*-\\
\\
Variables:\\
!<INPUT 0>! -- Demonstration\\
!<INPUT 1>! -- Context\\
!<INPUT 2>! -- Question\\
!<INPUT 3>! -- Options\\
\\
<commentblockmarker>\#\#\#</commentblockmarker>\\
\\
\#\#\# Demonstration\\
!<INPUT 0>!\\
\\
\#\#\# Context:\\
!<INPUT 1>!\\
\\
\#\#\# Question:\\
!<INPUT 2>!\\
\\
\#\#\# Options:\\
!<INPUT 3>!\\
\\
Your output will contain the following:\\
\#\#\# Thought: please output your thinking process step by step.\\
\#\#\# Evidence: Please output what you think is the Evidence in the topic. Please note that the Evidence needs to be directly from the question, i.e. it is the original string in the question, which can be matched directly to the original text by string matching.\\
\#\#\# Answer: When generating answers, please output only the serial numbers of the options.\\
\\
Please follow the format below for your output:\\
\\
\#\#\# Thought:\\
xxxxx\\
\\
\#\#\# Evidence:\\
xxxxx\\
\\
\#\#\# Answer:\\
1/2/3/4\\
    \hline
    \end{tabular}}
    \caption{Prompt of \texttt{Auto-CoT} method}
    \label{tab:auto_cot}
    
\end{table*}\begin{table*}[!ht]
    \centering
    \resizebox{\textwidth}{!}{
    \begin{tabular}{|p{\textwidth}|}
    \hline
\# -*- coding: utf-8 -*-\\
\\
Variables:\\
!<INPUT 0>! -- Context\\
!<INPUT 1>! -- Question\\
!<INPUT 2>! -- Options\\
\\
<commentblockmarker>\#\#\#</commentblockmarker>\\
\\
Below I will give you a detective reasoning question, please generate your thought process step by step based on the Context and the options and choose the answer you think is correct.\\
Note: When generating the answer, please output only the serial number of the option.\\
\\
\#\#\# Context:\\
!<INPUT 0>!\\
\\
\#\#\# Question:\\
!<INPUT 1>!\\
\\
\#\#\# Options:\\
!<INPUT 2>!\\
\\
Your output will contain the following:\\
\#\#\# Thought: please generate 5 completely different perspectives of your reflections based on the questions and options.\\
\#\#\# Summary: Please output a summary of all your thinking.\\
\#\#\# Evidence: Please output what you think is the Evidence in the Context. Please note that the Evidence needs to be directly from the Context, i.e. it is the original string in the Context, which can be matched directly to the original text by string matching.\\
\#\#\# Answer: please output only the serial numbers.\\
\\
Please have your output follow the format below:\\
\\
\#\#\# Thought:\\
1. xxxxxx\\
2. xxxxxx\\
3. xxxxxx\\
4. xxxxxx\\
5. xxxxxx\\
\\
\#\#\# Summarize:\\
xxxxxx\\
\\
\#\#\# Evidence:\\
xxxxx\\
\\
\#\#\# Answers:\\
1/2/3/4\\
    \hline
    \end{tabular}}
    \caption{Prompt of \texttt{Self Consistency} method}
    \label{tab:selfconsistency}
\end{table*}

\begin{table*}[!ht]
    \centering
    \resizebox{\textwidth}{!}{
    \begin{tabular}{|p{\textwidth}|}
    \hline
\# -*- coding: utf-8 -*-\\
\\
Variables:\\
!<INPUT 0>! -- Context\\
!<INPUT 1>! -- Question\\
!<INPUT 2>! -- Options\\
!<INPUT 3>! -- Longest Chain of Thought\\
\\
<commentblockmarker>\#\#\#</commentblockmarker>\\
\\
Below I will give you a detective reasoning question, please generate your thought process step by step based on the question and the options and choose the answer you think is correct.\\
Note: When generating the answer, please output only the serial number of the option.\\
\\
\#\#\# Context:\\
!<INPUT 0>!\\
\\
\#\#\# Question:\\
!<INPUT 1>!\\
\\
\#\#\# Options:\\
!<INPUT 2>!\\
\\
\#\#\# Chain of thought:\\
!<INPUT 3>!\\
\\
Your output will contain the following:
\#\#\# Evidence: Please output what you consider to be the Evidence in the topic. Please note that the Evidence needs to be directly from the topic, i.e. it is a string originally in the topic that can be matched directly to the original text by string matching.\\
\#\#\# Answer: please output only the serial numbers.\\
\\
Please follow the format below for your output:\\
\\
\#\#\# Evidence:\\
xxxxx\\
\\
\#\#\# Answer:\\
1/2/3/4\\
    \hline
    \end{tabular}}
    \caption{Prompt of \texttt{Complexity CoT} method}
    \label{tab:complexitycot}
\end{table*}

\begin{table*}[!ht]
    \centering
    \resizebox{\textwidth}{!}{
    \begin{tabular}{|p{\textwidth}|}
    \hline
\# -*- coding: utf-8 -*-\\

Variables:\\
!<INPUT 0>! -- Context\\
!<INPUT 1>! -- Question\\
!<INPUT 2>! -- Options\\
\\
<commentblockmarker>\#\#\#</commentblockmarker>\\
\\
Below I will give you a detective reasoning question, please generate your thought process step by step based on the Context and the options and choose the answer you think is correct.\\
Note: When generating the answer, please output only the serial number of the option.\\
\\
\#\#\# Context:\\
!<INPUT 0>!\\
\\
\#\#\# Question:\\
!<INPUT 1>!\\
\\
\#\#\# Options:\\
!<INPUT 2>!\\
\\
Your output will contain the following:\\
\#\#\# Thought: Please start with a general plan of how you intend to deal with the problem, and then think step-by-step about how to solve it based on your plan.\\
\#\#\# Evidence: please output what you think is the Evidence in the Context. Please note that the Evidence needs to be directly from the Context, i.e. it is the original string in the Context, which can be matched directly to the original text by string matching.\\
\#\#\# Answer: please output only the serial numbers.\\
\\
Please have your output follow the format below:\\
\\
\#\#\# Thought:\\
xxxxxx\\
\\
\#\#\# Evidence:\\
xxxxx\\
\\
\#\#\# Answer:\\
1/2/3/4\\
    \hline
    \end{tabular}}
    \caption{Prompt of \texttt{Plan and Solve CoT} method}
    \label{tab:pscot}
\end{table*}

\begin{table*}[!ht]
    \centering
    \resizebox{\textwidth}{!}{
    \begin{tabular}{|p{\textwidth}|}
    \hline
\# -*- coding: utf-8 -*-\\
\\
Variables:\\
!<INPUT 0>! -- Context\\
!<INPUT 1>! -- Question\\
!<INPUT 2>! -- Options\\
\\
<commentblockmarker>\#\#\#</commentblockmarker>\\
\\
Below I will give you a detective reasoning question, please generate your thought process step by step based on the Context and the options and choose the answer you think is correct.\\
Note: When generating the answer, please output only the serial number of the option.\\
\\
\#\#\# Context:\\
! <INPUT 0>!\\
\\
\#\#\# Question:\\
! <INPUT 1>!\\
\\
\#\#\# Options:\\
! <INPUT 2>!\\
\\
Your output will contain the following:\\
\#\#\# Clues: Feel free to summarize all possible clues in the Context\\
\#\#\# Connection: Feel free to correlate the clues you summarized above and introduce new clues that may exist.\\
\#\#\# Thought: Feel free to reason and think deeply about the clues you have summarized in the two steps above.\\
\#\#\# Summarize: Summarize all the thinking from the perspective of solving the problem in the Context.\\
\#\#\# Evidence: Please output what you think is the Evidence in the Context. Please note that the Evidence needs to be the direct content of the Context, i.e. it is the original string in the Context, which can be matched directly to the original text by string matching.\\
\#\#\# Answer: Please output only the serial number.\\
\\
Please have your output follow the format below:\\
\\
\#\#\# Clues:\\
xxxxxx\\
\\
\#\#\# Connection:\\
xxxxxx\\
\\
\#\#\# Thought:\\
xxxxxx\\
\\
\#\#\# Summarize:\\
xxxxxx\\
\\
\#\#\# Evidence:\\
xxxxx\\
\\
\#\#\# Answer:\\
1/2/3/4\\
    \hline
    \end{tabular}}
    \caption{Prompt of \texttt{Detective Reasoning} method}
    \label{tab:selfquestion}
\end{table*}

\end{CJK}
\end{document}